\pdfoutput=1

\documentclass[11pt]{article}

\usepackage[]{acl}

\usepackage{times}
\usepackage{latexsym}
\usepackage[T1]{fontenc}
\usepackage[utf8]{inputenc}
\usepackage{microtype}
\usepackage{inconsolata}
\usepackage{graphicx}

\usepackage{CJKutf8}   
\usepackage{booktabs}   
\usepackage{tabularx}  
\usepackage{xcolor}    
\usepackage{colortbl}  
\usepackage{amssymb}  
\usepackage{wasysym}   
\usepackage{xspace}    
\usepackage{enumitem}  
\usepackage{url}       
\usepackage{pifont}
\newcommand{\gcheck}{{\color{green!55!black}\ding{51}}}
\newcommand{\rx}{{\color{red!75!black}\ding{55}}}

\newcommand{\dataname}{\textsc{PILAPS}\xspace}

\definecolor{gray-light}{HTML}{e0e0e0}
\definecolor{gray-blue}{HTML}{EAF2FB}
\definecolor{blt}{HTML}{3d5a80}
\definecolor{bltl}{HTML}{98c1d9}
\definecolor{orange}{HTML}{ee6c4d}
\definecolor{dred}{HTML}{d90429}

\newcommand{\gbc}{\cellcolor{gray-blue}}
\newcommand{\gbl}{\cellcolor{gray-light}}

\newcolumntype{Y}{>{\raggedright\arraybackslash}X}

\newenvironment{promptbox}
  {\begin{list}{}{%
     \setlength{\leftmargin}{0.5pt}%
     \setlength{\rightmargin}{0.5pt}%
     \setlength{\labelwidth}{0pt}%
     \setlength{\labelsep}{0pt}%
     \setlength{\itemindent}{0pt}%
     \setlength{\listparindent}{0pt}%
     \setlength{\topsep}{6pt}%
     \setlength{\parsep}{5pt}}%
   \item\relax}
  {\end{list}}

\title{PILAPS: A Bilingual Benchmark for Evaluating Pragmatic Inference across Phenomena and Settings}

\author{Jiani Cao\textsuperscript{1} \and Anna Korhonen\textsuperscript{2} \and
  Chen Cecilia Liu\textsuperscript{2} \\
  \textsuperscript{1} Department of Theoretical and Applied Linguistics, University of Cambridge\\ 
  \textsuperscript{2} Language Technology Lab, University of Cambridge\\
  }

\begin{document}
\maketitle

\begin{abstract}
Recent work reports strong pragmatic performance by large language models (LLMs) in English, yet whether this performance generalises across languages and evaluation settings remains unclear. This question is increasingly important as LLMs are deployed across languages, cultures, and conversational contexts. In this paper, we first introduce PILAPS, a bilingual English/Mandarin Chinese benchmark covering eight pragmatic phenomena across purpose-written and dialogue-derived items, with both culturally adapted and independently sourced data. We then evaluate eight LLMs from four families across a range of model sizes and prompting strategies. Our results reveal substantial variation across models and pragmatic phenomena: open-source models continue to struggle particularly with humour, while cross-linguistic performance varies with model origin. Our error analysis shows that, compared with human participants, models are more prone to literal interpretations in conversational contexts. Theory-of-Mind prompting lowers the literal share of errors, whereas Chain-of-Thought prompting slightly raises it. Together, these findings demonstrate that strong pragmatic performance in one evaluation setting does not necessarily generalise across languages or conversational contexts.\footnote{\url{https://github.com/JianiCao/PILAPS}}

\end{abstract}

\section{Introduction}
\label{sec:intro}

\begin{figure}[h!]
\centering
\includegraphics[width=0.9\linewidth]{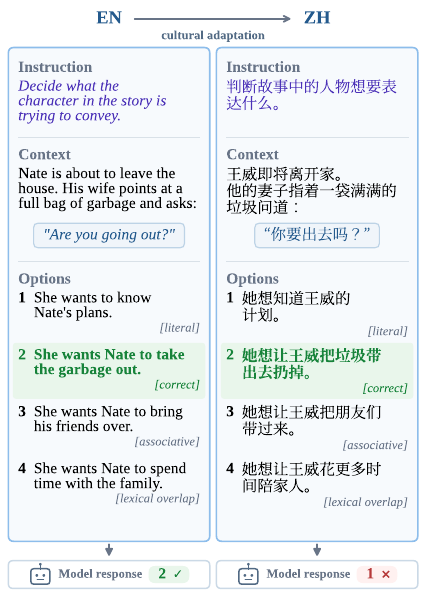}
\caption{Example items from \dataname{} (constructed component) illustrating cultural adaptation, distractor types, and model outputs. \dataname{} additionally includes a dialogue-derived component based on brief and extended dialogue.}
\label{fig:teaser}
\end{figure}

Understanding language requires recovering not only what is said but also what speakers intend beyond the literal meaning of an utterance \citep{grice1975logic, levinson1983pragmatics}. \textit{The kitchen is a bit warm} may function as a request to open a window, and an enthusiastic \textit{Wonderful} may express reluctant acceptance rather than genuine praise. This ability to infer intended meaning, known as pragmatic competence, is fundamental to conversational language understanding \citep{sperber1986relevance, clark1996using}. As large language models (LLMs) become increasingly integrated into conversational systems, recovering intended meaning has become essential for reliable interaction. A model that interprets an indirect refusal literally may comply with a rejected request, while one that treats a complaint as a factual statement may provide information instead of apologising. In both cases, the response may appear appropriate despite the model misjudging the speaker's intent.

Recent benchmarks and evaluation studies report strong pragmatic performance by frontier LLMs, particularly in English \citep{hu-etal-2023-fine, sravanthi-etal-2024-pub, strachan-etal-2024-testing}, yet whether this performance generalises beyond such settings remains unclear. The question matters as LLMs are increasingly deployed across languages, cultures, conversational environments, and model development contexts that differ substantially from those represented in English-centric evaluations. Robust pragmatic performance should not depend on a particular evaluation setup, but generalise across data types, languages, phenomena, and prompting strategies.

Testing such generalisation remains difficult because few benchmarks span both purpose-written and dialogue-derived materials, limiting evaluation across data types \citep{ma-etal-2025-pragmatics}. Moreover, because pragmatic interpretation hinges on language-specific idioms, wordplay, and cultural knowledge, cross-linguistic differences on directly translated materials may reflect translation artefacts rather than model deficits \citep{liu-etal-2024-multilingual, park-etal-2025-fluid, brglez-vintar-2026-polyester}. Robust evaluation therefore demands resources that combine varied pragmatic data types with careful cross-linguistic adaptation.

To address these challenges, we introduce \textbf{\dataname{}} (\textbf{P}ragmatic \textbf{I}nference across \textbf{LA}nguages, \textbf{P}henomena and \textbf{S}ettings), a bilingual benchmark (English/Mandarin Chinese) of 1{,}886 multiple-choice items covering eight pragmatic phenomena. \dataname{} comprises a \textbf{constructed} component covering seven phenomena and a \textbf{dialogue-derived} component covering four, spanning brief and extended dialogue. The constructed and brief English sets have culturally adapted Chinese counterparts, whereas the extended set is independently sourced in each language, with native-speaker baselines collected for that set. Figure~\ref{fig:teaser} shows a representative bilingual item from the constructed component.

Evaluation of eight instruction-tuned models across diverse scales, families, and prompting strategies    reveals substantial variation in pragmatic performance across tasks and languages, a decline from constructed to extended dialogue items, and cross-linguistic differences associated with model origin.\footnote{We categorise models by the primary language context in which they were developed: Chinese-origin (e.g., Qwen, DeepSeek) and English-origin (e.g., Llama, Gemini). We refer to this as \emph{model origin} throughout. This distinction is inherently entangled with model families, and is treated descriptively throughout.} Reasoning prompts provide limited and setting-dependent benefits.

In summary, we (1) introduce \dataname{}, a bilingual evaluation benchmark covering eight pragmatic phenomena across constructed and dialogue-derived materials; (2) develop a cultural adaptation protocol for cross-linguistic pragmatic evaluation; and (3) show that pragmatic performance does not generalise uniformly across evaluation settings, languages, and prompting strategies, with English-only evaluation concealing systematic cross-linguistic differences.

\section{Related Work}
\label{sec:related}

\begin{table*}[t]
\centering
\small
\resizebox{0.92\linewidth}{!}{
\begin{tabular}{@{}lcccccc@{}}
\toprule
& & & \multicolumn{2}{c}{\textbf{Source}} & \multicolumn{2}{c}{\textbf{Cross-lingual}} \\
\cmidrule(lr){4-5} \cmidrule(lr){6-7}
\textbf{Benchmark} & \textbf{Lang.} & \textbf{Phen.}
& Constructed & Derived & Adapted & Native \\
\midrule
PragMega \citep{floyd-etal-2023-pragmega}              & 1  & 7 & \gcheck & \rx     & --      & -- \\
MultiPragEval \citep{park-etal-2024-multiprageval}     & 4  & 4 & \gcheck & \rx     & \gcheck & \rx \\
PUB \citep{sravanthi-etal-2024-pub}                    & 1  & 4 & \gcheck & \rx     & --      & -- \\
SwordsmanImp \citep{yue-etal-2024-large}               & 1  & 1 & \rx     & \gcheck & --      & -- \\
FLUID QA \citep{park-etal-2025-fluid}                  & 3  & 4 & \gcheck & \rx     & \gcheck & \rx \\
MIDI \citep{almheiri-etal-2026-multilingual}           & 18 & 1 & \gcheck & \rx     & \rx     & \gcheck \\
VakyArth \citep{singh-etal-2026-vakyarth}              & 4  & 5 & \gcheck & \rx     & \rx     & \gcheck \\
\midrule
\dataname{}                                            & 2  & 8 & \gcheck & \gcheck & \gcheck & \gcheck \\
\bottomrule
\end{tabular}}
\caption{Comparison of recent pragmatic benchmarks. Lang: number of languages; Phen: phenomena covered; Source: whether items are purpose-written for evaluation (Constructed) or derived from pre-existing dialogue such as film or TV scripts (Derived); Cross-lingual: whether non-English items are culturally adapted or natively sourced. `--': monolingual.}
\label{tab:benchmarks}
\end{table*}

A growing number of benchmarks evaluate pragmatic reasoning in LLMs \citep{ma-etal-2025-pragmatics}, varying in source data, item construction, and target phenomena. Constructed benchmarks use purpose-written items targeting specific phenomena \citep{hu-etal-2023-fine, sravanthi-etal-2024-pub}. While they offer greater experimental control and tractability, they typically rely on brief items that capture limited aspects of real-world conversational usage. Dialogue-derived benchmarks instead draw from film and television scripts \citep{castro-etal-2019-towards, george-mamidi-2020-conversational, yue-etal-2024-large}. These benchmarks provide richer conversational context but less systematic control over the phenomena represented. The importance of contextualised evaluation is also reflected in model performance: while frontier models achieve high accuracy on constructed items, they continue to lag behind human performance in communicative settings \citep{qiu-etal-2025-wavelength}.

Despite recent progress towards multilingual evaluation, pragmatic benchmarking remains predominantly English-focused \citep{ma-etal-2025-pragmatics, singh-etal-2026-vakyarth}. Many benchmarks in languages other than English are monolingual \citep{yue-etal-2024-large, brglez-vintar-2026-polyester}, while multilingual benchmarks vary substantially in their construction (synthetic vs.\ human-authored), task formulation (identification vs.\ reasoning), and phenomenon coverage \citep{park-etal-2024-multiprageval, park-etal-2025-fluid, almheiri-etal-2026-multilingual}. These differences limit systematic cross-linguistic comparisons.

\dataname{} addresses these gaps by combining constructed and dialogue-derived resources under a common evaluation protocol, enabling comparisons across languages, pragmatic phenomena, and evaluation settings. As Table~\ref{tab:benchmarks} shows, it is the only benchmark among those compared that combines both source types with culturally adapted and natively sourced non-English data.

\section{\dataname{}}
\label{sec:benchmark}

\begin{figure*}[t]
\centering
\includegraphics[width=0.9\linewidth]{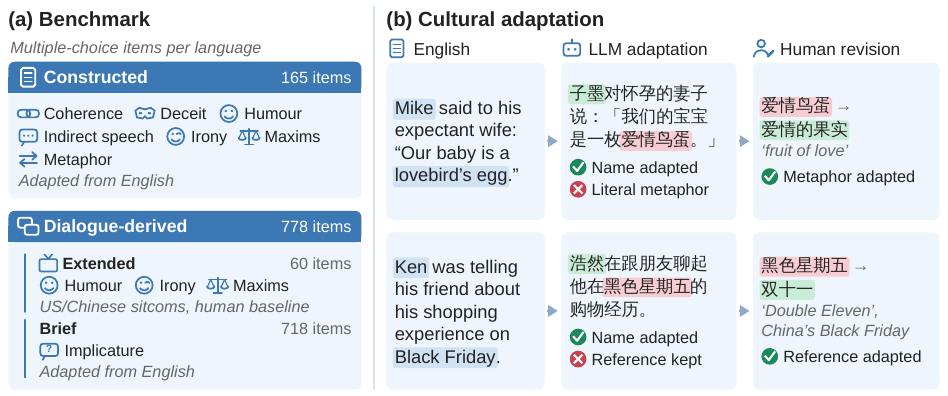}
\caption{Overview of \dataname{}. (a) Benchmark components, phenomena and items per language. (b) Cultural adaptation examples, with correctly adapted spans in green and unadapted spans in red. Full details of the adaptation procedure are given in Appendix~\ref{app:cultural}.}
\label{fig:overview}
\end{figure*}

\dataname{} is a bilingual multiple-choice benchmark for evaluating pragmatic inference in English and Chinese. It contains 1{,}886 items (943 per language, 2--5 options each) across eight pragmatic phenomena. The benchmark comprises a \textbf{constructed} component of purpose-written items and a \textbf{dialogue-derived} component built from pre-existing dialogue, with the latter including a \textbf{brief} set and an \textbf{extended} set. The constructed and brief English sets draw from existing evaluation resources, with matched Chinese versions produced through cultural adaptation, whereas the extended set is developed independently from source dialogue in each language. Figure~\ref{fig:overview} and Table~\ref{tab:phenomena} summarise the statistics and tasks of the benchmark.

\begin{table*}[t]
\centering
\small
\resizebox{0.92\linewidth}{!}{
\setlength{\tabcolsep}{6pt}
\begin{tabularx}{\textwidth}{@{}l Y Y c@{}}
\toprule
\textbf{Phenomenon} & \textbf{Definition} & \textbf{What the task tests} & \textbf{Num. Options} \\
\midrule
Coherence & Sentences that fit together into a sensible story & Judging whether two sentences form a coherent story & 2 \\ 
Deceit & Misleading someone in order to achieve a hidden goal & Recognising an intention to mislead & 4 \\[3pt]
Humour & A joke ending that is surprising but still fitting & Identifying the intended humorous continuation & 5 \\ 
Indirect Speech & Expressing a request or intention without stating it outright & Recognising the intended meaning of an indirect utterance & 4 \\ 
Irony & Saying the opposite of what one actually means & Recognising that a speaker means the opposite of the literal utterance & 4 \\ 
Maxims & Breaking a conversational rule to imply extra meaning & Inferring meaning from a flouted Gricean maxim & 4 \\ 
Metaphor & Describing one thing as if it were another thing & Identifying the intended metaphorical meaning & 5 \\ 
Implicature & Suggesting a meaning without saying it directly & Inferring whether an indirect reply is affirmative or negative & 2 \\
\bottomrule
\end{tabularx}}
\caption{The eight pragmatic phenomena in \dataname{}, with the number of answer options per item. Example items in Appendix~\ref{app:examples}.}
\label{tab:phenomena}
\end{table*}

\subsection{Dataset Construction}
\label{subsec:construction}

\paragraph{Constructed Component.} The English data are drawn from PragMega \citep{floyd-etal-2023-pragmega}, with matched Chinese versions produced through the cultural adaptation procedure described below. Each item comprises a short written scenario ending in a target utterance and two to five candidate interpretations. Following the task formulation of \citet{hu-etal-2023-fine}, most items require selecting the best interpretation for the context, whereas Humour requires choosing the continuation that completes the joke. We exclude five English Humour items that rely on puns that could not be preserved in Chinese, yielding 165 matched items per language.

Each answer option carries a task-specific label from the source material, reflecting phenomenon-specific misinterpretations. We preserve these labels in the Chinese adaptation. For example, Deceit items include \textit{IncorrectLiteral} options, which take the utterance at face value, and \textit{IncorrectNonLiteralSocialConvention} options, which interpret it as a conventional politeness formula. Labels are detailed in Table~\ref{tab:labels} (Appendix~\ref{app:erroranalysis}) and guide our error analysis (\S\ref{subsec:analysis}).

\paragraph{Dialogue-derived Component.}
The extended set draws its English and Chinese dialogue from \textit{The Big Bang Theory} \citep{lorre-prady-2007-bbt} and \begin{CJK*}{UTF8}{gbsn}武林外传\end{CJK*} (\textit{My Own Swordsman}; \citealp{ning-2006-wulin}), respectively, with dialogue selected independently for the two languages.\footnote{English transcripts were obtained from \url{https://www.springfieldspringfield.co.uk/episode_scripts.php?tv-show=big-bang-theory}. Chinese dialogue selection was informed by the SwordsmanImp corpus \citep{yue-etal-2024-large}, supplemented by author-transcribed material from the original broadcast.} This set comprises 60 items per language covering Humour, Irony and Maxims, the three phenomena for which \citet{hu-etal-2023-fine} report the largest gaps between model and human performance. To reduce source-recognition effects, we condense multi-scene contexts into short scenarios and replace character names with generic substitutes. The dialogue wording itself is otherwise preserved from the source scripts. Each item contains four or five answer options written by the authors, except for Humour, where the correct continuation is the original punchline from the script. Distractors follow the same annotation scheme as the constructed component.

The brief set adapts LUDWIG \citep{george-mamidi-2020-conversational}, an English conversational implicature dataset, following the binary evaluation protocol of \citet{ruis-etal-2023-gold}. Each item pairs a question and an indirect reply with affirmative and negative candidate interpretations. Cultural adaptation into Chinese yields 718 parallel items per language.

\paragraph{Cultural Adaptation.} Literal translation can obscure idioms, wordplay, and cultural references, shifting evaluation from pragmatic interpretation to basic comprehension \citep{park-etal-2025-fluid, brglez-vintar-2026-polyester}. We therefore culturally adapt the constructed and brief items into Chinese.

We develop a two-stage cultural adaptation procedure combining LLM-assisted adaptation with human evaluation and revision. In the first stage, structured prompts guide LLMs to preserve the pragmatic function of each item while adapting culturally specific references, such as names, locations and idioms, where necessary. In the second stage, annotators perform manual validation and revision, focusing on linguistic and cultural appropriateness. Figure~\ref{fig:overview} shows representative examples of cultural adaptation. We use the same structured prompt across pragmatic phenomena except Metaphor, for which we add instructions grounded in conceptual metaphor theory \citep{lakoff-johnson-1980-metaphors, lakoff-1993-contemporary} to preserve the underlying conceptual mapping while producing natural Chinese expressions. Full details of the procedure and prompts are provided in Appendix~\ref{app:cultural}.

Each item in the constructed component is adapted independently by three LLMs: GPT-4o \citep{openai2024gpt4osystemcard}, Claude Opus 4.5 \citep{anthropic2025claudeopus45} and Gemini 2.0 Flash \citep{googledeepmind2025gemini20flash}. Annotators score the three candidates on fluency, adequacy and compliance, each on a five-point scale, and the highest-scoring candidate is selected for revision. The full rubric is given in Appendix~\ref{appendix:quality}. 
The brief dialogue items are adapted with GPT-4o alone, given their volume and simpler task format. Each item has two polar-opposite answer choices, most commonly yes/no (see \citealp{ruis-etal-2023-gold}), making comparative selection across multiple adaptations unnecessary. None of the models used in dataset construction are included in the evaluation. Additional quality control for the brief and extended dialogue sets is described in Appendix~\ref{appendix:quality}. A human performance baseline is further established for the extended set, as outlined in \S\ref{subsec:evalsetup}.

\section{Experimental Setup}
\label{sec:setup}

\subsection{Evaluation}
\label{subsec:evalsetup}

\paragraph{Models.} We evaluate eight instruction-tuned LLMs from four families, varying in model origin and scale (Small: ${\leq}10$B parameters; Large: ${>}10$B or undisclosed). The four Chinese-origin models are DeepSeek-V4-Flash \citep{deepseekai2026deepseekv4}, Qwen3-32B, Qwen3-8B and Qwen3-4B-Instruct-2507 \citep{yang2025qwen3technicalreport}. The four English-origin models are Gemini-3.5-Flash \citep{googledeepmind2026gemini35flash}, Llama-3.3-70B, Llama-3-8B and Llama-3.2-3B \citep{grattafiori2024llama3herdmodels}. DeepSeek-V4-Flash and Gemini-3.5-Flash are accessed via API and the rest via Hugging Face \texttt{transformers} \citep{wolf-etal-2020-transformers}. Table~\ref{tab:models} in Appendix~\ref{app:exp_setup} gives full model details. 

\paragraph{Prompting Methods.} We evaluate models using three prompting methods. Since measured performance is sensitive to task phrasing \citep{mizrahi-etal-2024-state}, each method has three prompt variants that preserve the same strategy while varying in wording and specificity. Full prompts are given in Appendix~\ref{app:prompts}.
\begin{itemize}[nosep,leftmargin=1.2em]

\item Vanilla: Baseline prompting adapted from \citet{hu-etal-2023-fine}, querying the answer directly.
\item CoT: Chain-of-Thought prompting \citep{wei-etal-2022-chain}, eliciting general step-by-step reasoning before answering.
\item ToM: Theory-of-Mind prompting \citep{premack1978does, wilf-etal-2024-think}, eliciting reasoning about speaker mental states and intent before answering.
\end{itemize}

Both CoT and ToM follow a two-step protocol \citep{kojima-etal-2022-large}: Step 1 elicits reasoning given the item, and Step 2 presents the task instruction with the Step-1 response as context. All models are run with temperature 0 and a maximum of 1024 tokens. For reasoning models (Qwen3, DeepSeek-V4-Flash, Gemini-3.5-Flash), we disable or minimise native reasoning to reduce interference with the prompting manipulation.

\paragraph{Human Evaluation.} 
To assess pragmatic validity and establish a human performance baseline for the extended set, 10 native English and 10 native Mandarin Chinese speakers from the University of Cambridge evaluated all 60 items in their native language. Given the sample size, human results serve solely as descriptive baselines and are excluded from statistical analyses. The study received ethics approval from the University of Cambridge. Recruitment procedures and participant instructions
are detailed in Appendix~\ref{app:human_instructions}.

\subsection{Analysis Method}
\label{subsec:analysis}

\paragraph{Statistical Analysis.} We fit trial-level logistic regression models using generalised estimating equations (GEE; \citealp{liang-zeger-1986-longitudinal}) with item-level clustering to account for repeated evaluations of the same item across models and prompting conditions. Model scale and prompt variant are included as additive covariates. Predictor coding is detailed in Appendix~\ref{app:stats}. 

We test four effects: (1) phenomenon variation in constructed items (omitting binary-choice Coherence because its higher chance level makes raw accuracy less directly comparable with the other phenomena); (2) transfer from constructed to extended dialogue items across shared phenomena (Humour, Irony and Maxims); (3) cross-linguistic differences and their interaction with model origin, controlling for setting; and (4) interactions between prompting method and evaluation setting. The first three analyses use vanilla prompting only, treating CoT and ToM as interventions rather than measures of baseline pragmatic ability. The prompting analysis uses all three methods. For robustness, we re-estimate the cross-linguistic model across all responses, adjusting for prompting method, and perform leave-one-model-out checks for model-origin interactions (Appendix~\ref{app:sub:statistical}).

\begin{table*}[t]
\centering
\resizebox{0.98\linewidth}{!}{
\begin{tabular}{l ccccccc c cccc c c}
\toprule
& \multicolumn{8}{c}{\textbf{Constructed}} & \multicolumn{5}{c}{\textbf{Dialogue-derived}} & \\
\cmidrule(lr){2-9} \cmidrule(lr){10-14}
\textbf{Model}
  & Coh & Dec & Hum & Ind & Iro & Max & Met & \textit{Avg}
  & Hum & Iro & Max & Imp & \textit{Avg} & \textbf{Mean} \\
\midrule
\multicolumn{15}{c}{\gbl\textit{English Dataset}} \\
\midrule
DeepSeek-V4-Flash  & 80.83 & 93.33 & 93.33 & 93.33 & 80.00 & 85.00 & 90.00 & \gbc\textit{87.98} & 61.67 & 85.00 & 65.00 & 65.83 & \gbc\textit{69.37} & \textbf{81.21} \\
Qwen3-32B          & 79.17 & 90.00 & 73.33 & 80.00 & 77.33 & 73.33 & 85.00 & \gbc\textit{79.74} & 50.00 & 50.00 & 51.67 & 72.84 & \gbc\textit{56.13} & \textbf{71.15} \\
Qwen3-8B           & 71.67 & 65.00 & 56.67 & 85.00 & 52.00 & 56.67 & 76.67 & \gbc\textit{66.24} & 28.33 & 46.67 & 43.33 & 66.99 & \gbc\textit{46.33} & \textbf{59.00} \\
Qwen3-4B           & 79.17 & 71.67 & 45.00 & 80.00 & 49.33 & 51.67 & 66.67 & \gbc\textit{63.36} & 40.00 & 40.00 & 43.33 & 75.07 & \gbc\textit{49.60} & \textbf{58.35} \\
\midrule
Gemini-3.5-Flash   & 91.67 & 100.00 & 100.00 & 96.67 & 97.33 & 98.33 & 98.33 & \gbc\textit{97.48} & 78.33 & 95.00 & 86.67 & 85.77 & \gbc\textit{86.44} & \textbf{93.46} \\
Llama-3.3-70B      & 80.83 & 90.00 & 63.33 & 95.00 & 65.33 & 75.00 & 78.33 & \gbc\textit{78.26} & 26.67 & 48.33 & 46.67 & 64.90 & \gbc\textit{46.64} & \textbf{66.76} \\
Llama-3-8B         & 58.33 & 78.33 & 35.00 & 75.00 & 53.33 & 43.33 & 65.00 & \gbc\textit{58.33} & 30.00 & 35.00 & 40.00 & 69.50 & \gbc\textit{43.62} & \textbf{52.98} \\
Llama-3.2-3B       & 52.50 & 78.33 & 23.33 & 88.33 & 50.67 & 63.33 & 45.00 & \gbc\textit{57.36} & 18.33 & 35.00 & 45.00 & 62.59 & \gbc\textit{40.23} & \textbf{51.13} \\
\midrule
\textit{Avg (cross-model)}      & 74.27 & 83.33 & 61.25 & 86.67 & 65.67 & 68.33 & 75.62 & \gbc\textit{73.59} & 41.67 & 54.38 & 52.71 & 70.44 & \gbc\textit{54.80} & \textbf{66.76} \\
\midrule
\multicolumn{15}{c}{\gbl\textit{Chinese Dataset}} \\
\midrule
DeepSeek-V4-Flash  & 90.00 & 95.00 & 88.33 & 98.33 & 81.33 & 85.00 & 88.33 & \gbc\textit{89.48} & 78.33 & 78.33 & 88.33 & 76.65 & \gbc\textit{80.41} & \textbf{86.18} \\
Qwen3-32B          & 78.33 & 90.00 & 73.33 & 90.00 & 76.00 & 75.00 & 90.00 & \gbc\textit{81.81} & 41.67 & 85.00 & 51.67 & 77.16 & \gbc\textit{63.87} & \textbf{75.29} \\
Qwen3-8B           & 77.50 & 70.00 & 65.00 & 90.00 & 53.33 & 55.00 & 90.00 & \gbc\textit{71.55} & 30.00 & 33.33 & 50.00 & 71.31 & \gbc\textit{46.16} & \textbf{62.32} \\
Qwen3-4B           & 80.00 & 75.00 & 40.00 & 73.33 & 60.00 & 50.00 & 88.33 & \gbc\textit{66.67} & 46.67 & 30.00 & 33.33 & 72.84 & \gbc\textit{45.71} & \textbf{59.05} \\
\midrule
Gemini-3.5-Flash   & 90.00 & 95.00 & 100.00 & 100.00 & 96.00 & 91.67 & 90.00 & \gbc\textit{94.67} & 85.00 & 95.00 & 90.00 & 88.85 & \gbc\textit{89.71} & \textbf{92.87} \\
Llama-3.3-70B      & 75.83 & 66.67 & 50.00 & 83.33 & 62.67 & 50.00 & 71.67 & \gbc\textit{65.74} & 10.00 & 35.00 & 23.33 & 69.08 & \gbc\textit{34.35} & \textbf{54.33} \\
Llama-3-8B         & 40.00 & 61.67 & 18.33 & 75.00 & 53.33 & 31.67 & 48.33 & \gbc\textit{46.90} & 30.00 & 20.00 & 26.67 & 56.13 & \gbc\textit{33.20} & \textbf{41.92} \\
Llama-3.2-3B       & 50.00 & 40.00 & 33.33 & 36.67 & 42.67 & 28.33 & 33.33 & \gbc\textit{37.76} & 16.67 & 41.67 & 28.33 & 47.77 & \gbc\textit{33.61} & \textbf{36.25} \\
\midrule
\textit{Avg (cross-model)}      & 72.71 & 74.17 & 58.54 & 80.83 & 65.67 & 58.33 & 75.00 & \gbc\textit{69.32} & 42.29 & 52.29 & 48.96 & 69.97 & \gbc\textit{53.38} & \textbf{63.52} \\
\bottomrule
\end{tabular}}
\caption{Accuracy (\%) on each task under vanilla prompting, averaged across the three prompt variants. Coh = Coherence, Dec = Deceit, Hum = Humour, Ind = Indirect Speech, Iro = Irony, Max = Maxims, Met = Metaphor, Imp = Implicature. Within the dialogue-derived component, Humour, Irony and Maxims are the extended tasks and Implicature the brief task. \textit{Avg} is the unweighted mean over the preceding block and \textbf{Mean} the unweighted mean over all eleven tasks. Within each block, the first four models are Chinese-origin and the last four English-origin.}
\label{tab:pragmatics}
\end{table*}

\paragraph{Error Analysis.} Beyond accuracy, we analyse model error types and, for the extended dialogue items, compare model and human error patterns. We group the task-specific labels described in \S\ref{subsec:construction} into four broader categories to enable cross-task comparisons. We apply these categories to the six phenomena with labelled distractors. Coherence and Implicature are excluded because their binary-choice tasks do not provide distractor labels. We distinguish (1) \textit{Literal}, responses that take the face-value reading when a non-literal interpretation is required; (2) \textit{Wrong-pragmatic}, responses that draw a non-literal inference but arrive at the wrong interpretation; (3) \textit{Associative}, responses driven by surface-level association with the scenario rather than by pragmatic inference; and (4) \textit{Non-sequitur}, responses that bear no meaningful relation to the speaker's communicative intent. The full mapping is in Appendix~\ref{app:erroranalysis}.

\section{Results and Discussion}
\label{sec:results}

\subsection{Main Results}
\label{subsec:main_results}
Table~\ref{tab:pragmatics} reports accuracy under vanilla prompting by model, task and evaluation language, averaged across the three prompt variants.
Overall, performance varies substantially across models and pragmatic phenomena. Among open-source models, task-level accuracy ranges from 10.00\% for Llama-3.3-70B on Chinese dialogue-derived Humour to 98.33\% for DeepSeek-V4-Flash on Chinese constructed Indirect Speech. Gemini-3.5-Flash achieves the highest overall accuracy in both languages. Smaller models (${\leq}10$B parameters) generally perform worse across tasks and languages. Within the constructed component, cross-model accuracy varies significantly across the six non-binary phenomena included in the analysis ($\chi^2(5) = 20.11$, $p = .001$), with Humour and Maxims among the more challenging phenomena.

\textbf{Extended dialogue items are significantly more difficult for models.} Averaged across Humour, Irony and Maxims, every model performs worse on the extended dialogue items than on the constructed items in both languages. The decline in accuracy remains significant after controlling for model origin, scale and prompt variant ($\beta = -0.655$, $\mathrm{SE} = 0.170$, $p < .001$), and does not differ significantly across the three phenomena or the two languages (setting $\times$ phenomenon $p = .79$, setting $\times$ language $p = .68$, three-way $p = .78$).

Most models also fall below native-speaker baselines across all three extended phenomena, though the gap size varies by model and phenomenon. Native speakers achieve 66.00/85.00/79.00 on English Humour/Irony/Maxims and 66.50/90.00/91.00 on their Chinese counterparts (see Appendix~\ref{app:human}). This suggests that the lower model accuracy on extended dialogue items cannot be attributed solely to their difficulty for human readers. Extended dialogue may require greater integration of information across turns and greater reliance on implicit social and linguistic cues, making pragmatic inference more challenging for LLMs.

\textbf{Cross-linguistic performance varies with model origin.} Averaged equally across all eleven tasks and models regardless of origin, overall accuracy on \dataname{} is 66.76\% in English and 63.52\% in Chinese. A trial-level GEE model controlling for model origin, scale, prompt variant and evaluation setting shows no significant aggregate language effect ($\beta = -0.055$, $\mathrm{SE} = 0.036$, $p = .13$). However, the coefficient is item-weighted and does not directly test the macro-average difference.

The pattern differs when model origin is considered. English-origin models drop from 66.08\% in English to 56.34\% in Chinese ($\beta = -0.295$, $p < .001$), whereas Chinese-origin models increase from 67.43\% in English to 70.71\% in Chinese, producing a significant origin-by-language interaction ($\beta = +0.501$, $p < .001$). The two groups do not differ significantly in English ($\beta = +0.003$, $p = .94$) but diverge in Chinese ($\beta = +0.505$, $p < .001$). The interaction remains significant when each model is held out in turn (Appendix~\ref{app:sub:statistical}), indicating that it is not driven by any single model.

This asymmetry may reflect differences in pre-training language composition, such as the approximately 8\% multilingual data mixture in Llama 3 models \citep{grattafiori2024llama3herdmodels} versus the heavy emphasis on Chinese tokens reported for the DeepSeek family \citep{deepseekai2024deepseekv2}. Such differences may contribute to variation in cross-lingual performance and cultural alignment \citep{blevins-zettlemoyer-2022-language, alkhamissi-etal-2024-investigating, kew-etal-2024-turning}, particularly for pragmatic inference, which depends on language- and culture-specific conventions beyond linguistic form \citep{ying-etal-2025-disentangling}.

\subsection{Error Analysis}
\label{subsec:error_analysis}

\begin{figure}[t]
    \centering
    \includegraphics[width=0.85\linewidth]{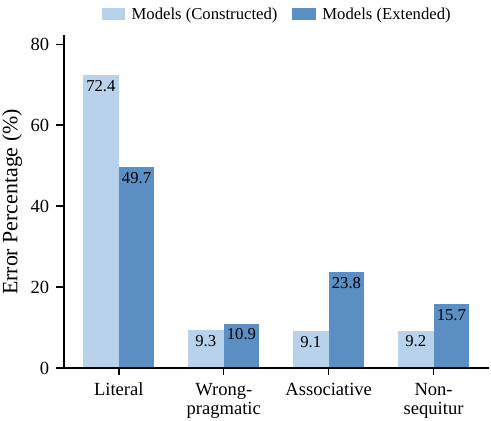}
    \caption{Error-family composition of model responses under vanilla prompting. Shares of classifiable incorrect responses are computed within each task, pooling the eight models, then averaged across the six constructed tasks or the three extended tasks, and across the two languages.}
    \label{fig:error_analysis}
\end{figure}

Overall, literal responses are the most prevalent error type (Figure~\ref{fig:error_analysis}), forming the largest error family in all but one of the 32 model-by-language cells covered by the error analysis (Table~\ref{tab:errormodel} in Appendix~\ref{app:erroranalysis}).

In the constructed component, literal responses account for 78.92\% of errors in English and 65.95\% in Chinese (Table~\ref{tab:errortask} in Appendix~\ref{app:erroranalysis}). This pattern is particularly pronounced for Indirect Speech and Irony, where more than 84\% of errors are literal in both languages, whereas Humour produces a more dispersed error profile.

\begin{table}[h!]
\centering
\small
\setlength{\tabcolsep}{3.5pt}
\begin{tabular}{@{}lllrrrr@{}}
\toprule
& \textbf{Task} & & \textbf{Lit.} & \textbf{W-prag.}
& \textbf{Assoc.} & \textbf{N-seq.} \\
\midrule
EN & Humour & Models & 37.50 & -- & 26.79 & 35.71 \\
   &        & Humans & 20.59 & -- & 20.59 & 58.82 \\
   & Irony  & Models & 54.34 & -- & 42.47 & 3.20 \\
   &        & Humans & 44.83 & -- & 51.72 & 3.45 \\
   & Maxims & Models & 65.20 & 24.67 & 10.13 & -- \\
   &        & Humans & 59.52 & 26.19 & 14.29 & -- \\
\midrule
ZH & Humour & Models & 29.60 & -- & 24.55 & 45.85 \\
   &        & Humans & 11.94 & -- & 14.93 & 73.13 \\
   & Irony  & Models & 75.55 & -- & 15.28 & 9.17 \\
   &        & Humans & 15.00 & -- & 50.00 & 35.00 \\
   & Maxims & Models & 35.92 & 40.82 & 23.27 & -- \\
   &        & Humans & 16.67 & 72.22 & 11.11 & -- \\
\bottomrule
\end{tabular}
\caption{Error-family composition (\%) for models and human participants on extended dialogue. Model results use vanilla prompting and pool all eight models. Lit. = Literal; W-prag. = Wrong-pragmatic; Assoc. = Associative; N-seq. = Non-sequitur; `--' = distractor family unavailable for that task.}
\label{tab:modelhumanerrors}
\end{table}

\textbf{On extended dialogue, model errors are more strongly skewed towards literal interpretations than human errors.} As Table~\ref{tab:modelhumanerrors} shows, models produce a higher proportion of literal errors than human participants across all three phenomena in both languages. Relative to the constructed component, associative errors also account for a larger share of model errors on extended dialogue.

Models' tendency to select literal options may stem from the strong semantic overlap of these options with the target utterance, which can make them particularly salient under ambiguity. This account is consistent with evidence that lexical overlap can influence LLM performance and that models can favour literal interpretations in non-literal language tasks \citep{stowe-etal-2022-impli, hu-etal-2023-fine, sanchez-bayona-agerri-2025-metaphor}.

\subsection{Prompting Effects}
\label{subsec:prompts_effect}

\textbf{CoT and ToM prompting have model-, language- and setting-dependent effects.} Table~\ref{tab:cot_tom} shows that neither CoT nor ToM consistently improves performance across models, and both show greater sensitivity to wording than vanilla prompting (statistical analysis in Appendix~\ref{app:sub:prompts_var_analysis}). CoT helps the two larger Llama models and Qwen3-32B but lowers accuracy for the two smaller Qwen models and Gemini-3.5-Flash. ToM yields more consistent gains, with improvements in both languages for Llama-3.3-70B, Llama-3-8B, Qwen3-32B and DeepSeek-V4-Flash, although Gemini-3.5-Flash declines under both reasoning prompts. Breaking these results down by evaluation setting reveals further differences (Table~\ref{tab:promptsetting}, Appendix~\ref{app:sub:prompts_var_analysis}).

\begin{table}[h!]
\centering
\resizebox{0.98\linewidth}{!}{
\begin{tabular}{l ccc}
\toprule
\textbf{Model} & \textbf{Vanilla} & \textbf{CoT} & \textbf{ToM} \\
\midrule
DeepSeek-V4-Flash & 81.21 / 86.18 & 82.87 / 83.61 & 84.90 / 86.61 \\
Qwen3-32B & 71.15 / 75.29 & 73.67 / 75.66 & 74.56 / 78.77 \\
Qwen3-8B & 59.00 / 62.32 & 58.05 / 59.27 & 58.21 / 67.18 \\
Qwen3-4B & 58.35 / 59.05 & 57.12 / 56.75 & 59.60 / 58.23 \\
\midrule
Gemini-3.5-Flash & 93.46 / 92.87 & 93.33 / 92.22 & 91.82 / 91.45 \\
Llama-3.3-70B & 66.76 / 54.33 & 67.39 / 57.48 & 70.56 / 60.82 \\
Llama-3-8B & 52.98 / 41.92 & 57.85 / 47.20 & 57.32 / 48.95 \\
Llama-3.2-3B & 51.13 / 36.25 & 49.83 / 38.32 & 53.78 / 35.21 \\
\bottomrule
\end{tabular}}
\caption{Aggregated performance comparison across prompting strategies. Each cell reports mean accuracy over the eleven tasks, weighting tasks equally, and over the three prompt variants, as English / Chinese. The first four models are Chinese-origin and the last four English-origin.}
\label{tab:cot_tom}
\end{table}

Across English and Chinese, CoT produces its largest aggregated gain on the brief task (Implicature, $+4.53$ percentage points, interaction $\beta = +0.247$, $\mathrm{SE} = 0.065$, $p < .001$; Table~\ref{tab:promptsetting}), whereas ToM's corresponding gain in extended dialogue ($+4.58$ points) is not reliably larger than in other settings ($\beta = +0.083$, $p = .24$). The CoT gain on brief Implicature suggests that explicit intermediate reasoning may help models infer an affirmative or negative stance from short, indirect replies, consistent with prior reports that CoT prompting, including maxim-guided variants, improves binary implicature resolution \citep{kim-etal-2023-pope, ruis-etal-2023-gold}. The ToM gain on extended dialogue may reflect the importance of speaker perspective when relevant contextual information is distributed across multiple turns \citep{wilf-etal-2024-think, shinoda-etal-2026-lets}. Prompt effectiveness may thus reflect what an evaluation leaves implicit rather than an inherent benefit of extra reasoning.

\textbf{ToM reduces the proportion of literal errors, whereas CoT increases it.} The direction holds in all four setting-by-language comparisons covered by the error analysis (Appendix~\ref{app:erroranalysis}). The shift is particularly pronounced in the constructed component, where errors under ToM move away from literal responses towards wrong-pragmatic, associative and non-sequitur interpretations. This suggests that avoiding an unintended literal interpretation is distinct from successfully recovering the intended pragmatic meaning, consistent with evidence that apparent mental-state reasoning can rely on relatively shallow heuristics \citep{shapira-etal-2024-clever}.

\subsection{Additional Evaluation of Small Models}
\label{subsec:small_models}
Our main evaluation (Table~\ref{tab:pragmatics}) shows that smaller LLMs generally perform worse on pragmatic tasks. Given growing interest in deploying smaller LLMs in resource-constrained and agentic settings \citep{belcak2025smalllanguagemodelsfuture, miranda2026multilingual}, we additionally evaluate four 3--4B-parameter LLMs. These models generally struggle with pragmatic inference in both languages, particularly on extended dialogue, with Humour remaining the most challenging phenomenon on average (Appendix~\ref{app:sub:small_llm}).

\section{Conclusion}
\label{sec:conclusion}

We introduce \dataname{}, a bilingual benchmark for evaluating pragmatic inference across languages, phenomena, evaluation settings, and prompting strategies, combining constructed and dialogue-derived items with culturally adapted and independently sourced Chinese materials.

Our findings show that (1) models perform considerably better on constructed items than on extended dialogue items; (2) cross-linguistic performance varies with model origin; (3) model errors are more strongly skewed towards literal interpretations than human errors on extended dialogue, with ToM prompting reducing and CoT prompting increasing this tendency; and (4) the utility of reasoning prompts is model- and setting-dependent.

Together, these results show that strong pragmatic performance in one evaluation setting does not necessarily generalise across languages or conversational contexts.

\section*{Limitations}

\dataname{} covers English and Mandarin Chinese, both high-resource languages, leaving generalisation to low-resource languages uncertain. The extended dialogue items are drawn from a single TV series per language, so programme-specific variation cannot be fully separated from language. However, culturally adapted constructed items show similar cross-linguistic patterns, suggesting that programme-specific variation is unlikely to explain the full pattern.

Our analysis includes four Chinese-origin and four English-origin models, but model origin is confounded with model-family-specific factors in this sample. We therefore interpret the observed origin-by-language interaction as a \emph{descriptive} association rather than evidence for an independent effect of model origin. Broader coverage across model families and more detailed training documentation are needed to disentangle the factors underlying this pattern.

All \dataname{} items use a multiple-choice format, which restricts the space of possible interpretations and may not reflect models' ability to recover intended meaning independently. The error analysis likewise captures response outcomes rather than underlying mechanisms. Future work should extend evaluation to interactive settings requiring models to infer speaker meaning and generate contextually appropriate responses.

\section*{Acknowledgements}
This work was supported by the UK Research and Innovation (UKRI) Frontier Research Grant EP/Y031350/1 EQUATE.

LLMs were used to improve the clarity and fluency of the writing and to refine the phrasing. Coding agents were also used to assist with the development of the code for this work.

\bibliography{custom}

@article{kim-etal-2023-pope,
      title={``Is the Pope Catholic?'' Applying Chain-of-Thought Reasoning to Understanding Conversational Implicatures},
      author={Kim, Zae Myung and Taylor, David E. and Kang, Dongyeop},
      year={2023},
      eprint={2305.13826},
      journal      = {arXiv},
      volume       = {arXiv:2305.13826},
      archivePrefix={arXiv},
      primaryClass={cs.CL},
      url={https://arxiv.org/abs/2305.13826}
}

@inproceedings{almheiri-etal-2026-multilingual,
    title = "Multilingual Idioms in Sentences and Conversations Across High-, Medium-, and Low-Resource Languages",
    author = "Almheiri, Saeed  and
      Elbouardi, Bilal  and
      Pranida, Salsabila Zahirah  and
      Nikishina, Irina  and
      B, Ashwath Rao  and
      Krishnamurthy, Parameswari  and
      Airlangga, Muhammad Cendekia  and
      Genadi, Rifo Ahmad  and
      Bao, Nguyen Phan Gia  and
      Yari, Amir Hossein  and
      Toyin, Hawau Olamide  and
      Mukhituly, Nurdaulet  and
      Attia, Mena  and
      Hassan, Besher  and
      Hidayatullah, Ahmad Fathan  and
      Kuribayashi, Tatsuki  and
      Li, Haonan  and
      Bhat, Suma  and
      Koto, Fajri",
    editor = "Liakata, Maria  and
      Moreira, Viviane P.  and
      Zhang, Jiajun  and
      Jurgens, David",
    booktitle = "Proceedings of the 64th Annual Meeting of the {A}ssociation for {C}omputational {L}inguistics (Volume 1: Long Papers)",
    month = jul,
    year = "2026",
    address = "San Diego, California, United States",
    publisher = "Association for Computational Linguistics",
    url = "https://aclanthology.org/2026.acl-long.564/",
    doi = "10.18653/v1/2026.acl-long.564",
    pages = "12363--12389",
    ISBN = "979-8-89176-390-6",
}

@article{singh-etal-2026-vakyarth,
      title={VakyArth: Evaluating Pragmatic Competence in {LLM}s across {I}ndic Languages}, 
      author={Singh, Usneek and Veera Balaji Kumar, Poorvaja and Nanda, Parth and Madhusoodanan, Anand and Guo, Geyang and Xu, Wei and Li, Junyi Jessy},
      year={2026},
      eprint={2609.01788},
            journal      = {arXiv},
      volume       = {arXiv:2609.01788},
      archivePrefix={arXiv},
      primaryClass={cs.CL},
      url={https://arxiv.org/abs/2609.01788}
}

@article{moghaddam-honey-2023-boosting,
      title={Boosting Theory-of-Mind Performance in Large Language Models via Prompting}, 
      author={Rahimi Moghaddam, Shima and Honey, Christopher J.},
      year={2023},
      eprint={2304.11490},
      journal      = {arXiv},
      volume       = {arXiv:2304.11490},
      archivePrefix={arXiv},
      primaryClass={cs.AI},
      url={https://arxiv.org/abs/2304.11490}
}

@inproceedings{blevins-zettlemoyer-2022-language,
    title = "Language Contamination Helps Explain the Cross-lingual Capabilities of {E}nglish Pretrained Models",
    author = "Blevins, Terra  and
      Zettlemoyer, Luke",
    editor = "Goldberg, Yoav  and
      Kozareva, Zornitsa  and
      Zhang, Yue",
    booktitle = "Proceedings of the 2022 Conference on Empirical Methods in Natural Language Processing",
    month = dec,
    year = "2022",
    address = "Abu Dhabi, United Arab Emirates",
    publisher = "Association for Computational Linguistics",
    url = "https://aclanthology.org/2022.emnlp-main.233/",
    doi = "10.18653/v1/2022.emnlp-main.233",
    pages = "3563--3574",
}

@inproceedings{kew-etal-2024-turning,
    title = "Turning {E}nglish-centric {LLM}s Into Polyglots: How Much Multilinguality Is Needed?",
    author = "Kew, Tannon  and
      Schottmann, Florian  and
      Sennrich, Rico",
    editor = "Al-Onaizan, Yaser  and
      Bansal, Mohit  and
      Chen, Yun-Nung",
    booktitle = "Findings of the Association for Computational Linguistics: EMNLP 2024",
    month = nov,
    year = "2024",
    address = "Miami, Florida, USA",
    publisher = "Association for Computational Linguistics",
    url = "https://aclanthology.org/2024.findings-emnlp.766/",
    doi = "10.18653/v1/2024.findings-emnlp.766",
    pages = "13097--13124"
}

@inproceedings{alkhamissi-etal-2024-investigating,
    title = "Investigating Cultural Alignment of Large Language Models",
    author = "AlKhamissi, Badr  and
      ElNokrashy, Muhammad  and
      Alkhamissi, Mai  and
      Diab, Mona",
    editor = "Ku, Lun-Wei  and
      Martins, Andre  and
      Srikumar, Vivek",
    booktitle = "Proceedings of the 62nd Annual Meeting of the Association for Computational Linguistics (Volume 1: Long Papers)",
    month = aug,
    year = "2024",
    address = "Bangkok, Thailand",
    publisher = "Association for Computational Linguistics",
    url = "https://aclanthology.org/2024.acl-long.671/",
    doi = "10.18653/v1/2024.acl-long.671",
    pages = "12404--12422"
}

@article{deepseekai2024deepseekv2,
      title={DeepSeek-V2: A Strong, Economical, and Efficient Mixture-of-Experts Language Model}, 
      author={{DeepSeek-AI}},
      year={2024},
      eprint={2405.04434},
            journal      = {arXiv},
      volume       = {arXiv:2405.04434},
      archivePrefix={arXiv},
      primaryClass={cs.CL},
      url={https://arxiv.org/abs/2405.04434}
}

@inproceedings{sanchez-bayona-agerri-2025-metaphor,
    title = "Metaphor and Large Language Models: When Surface Features Matter More than Deep Understanding",
    author = "Sanchez-Bayona, Elisa  and
      Agerri, Rodrigo",
    editor = "Che, Wanxiang  and
      Nabende, Joyce  and
      Shutova, Ekaterina  and
      Pilehvar, Mohammad Taher",
    booktitle = "Findings of the Association for Computational Linguistics: ACL 2025",
    month = jul,
    year = "2025",
    address = "Vienna, Austria",
    publisher = "Association for Computational Linguistics",
    url = "https://aclanthology.org/2025.findings-acl.898/",
    doi = "10.18653/v1/2025.findings-acl.898",
    pages = "17462--17477",
    ISBN = "979-8-89176-256-5"
}

@inproceedings{stowe-etal-2022-impli,
    title = "{IMPLI}: Investigating {NLI} Models' Performance on Figurative Language",
    author = "Stowe, Kevin  and
      Utama, Prasetya  and
      Gurevych, Iryna",
    editor = "Muresan, Smaranda  and
      Nakov, Preslav  and
      Villavicencio, Aline",
    booktitle = "Proceedings of the 60th Annual Meeting of the Association for Computational Linguistics (Volume 1: Long Papers)",
    month = may,
    year = "2022",
    address = "Dublin, Ireland",
    publisher = "Association for Computational Linguistics",
    url = "https://aclanthology.org/2022.acl-long.369/",
    doi = "10.18653/v1/2022.acl-long.369",
    pages = "5375--5388",
}

@inproceedings{shinoda-etal-2026-lets,
    title = "Let{'}s Put Ourselves in Sally{'}s Shoes: Shoes-of-Others Prefilling Improves Theory of Mind in Large Language Models",
    author = "Shinoda, Kazutoshi  and
      Hojo, Nobukatsu  and
      Nishida, Kyosuke  and
      Yamazaki, Yoshihiro  and
      Suzuki, Keita  and
      Sugiyama, Hiroaki  and
      Saito, Kuniko",
    editor = "Demberg, Vera  and
      Inui, Kentaro  and
      Marquez, Llu{\'i}s",
    booktitle = "Findings of the {A}ssociation for {C}omputational {L}inguistics: {EACL} 2026",
    month = mar,
    year = "2026",
    address = "Rabat, Morocco",
    publisher = "Association for Computational Linguistics",
    url = "https://aclanthology.org/2026.findings-eacl.6/",
    doi = "10.18653/v1/2026.findings-eacl.6",
    pages = "95--109",
    ISBN = "979-8-89176-386-9",
}

@inproceedings{ying-etal-2025-disentangling,
    title = "Disentangling Language and Culture for Evaluating Multilingual Large Language Models",
    author = "Ying, Jiahao  and
      Tang, Wei  and
      Zhao, Yiran  and
      Cao, Yixin  and
      Rong, Yu  and
      Zhang, Wenxuan",
    editor = "Che, Wanxiang  and
      Nabende, Joyce  and
      Shutova, Ekaterina  and
      Pilehvar, Mohammad Taher",
    booktitle = "Proceedings of the 63rd Annual Meeting of the Association for Computational Linguistics (Volume 1: Long Papers)",
    month = jul,
    year = "2025",
    address = "Vienna, Austria",
    publisher = "Association for Computational Linguistics",
    url = "https://aclanthology.org/2025.acl-long.1082/",
    doi = "10.18653/v1/2025.acl-long.1082",
    pages = "22230--22251",
    ISBN = "979-8-89176-251-0"
}

@inproceedings{qiu-etal-2025-wavelength,
    title = "On the Same Wavelength? Evaluating Pragmatic Reasoning in Language Models across Broad Concepts",
    author = "Qiu, Linlu  and
      Zhang, Cedegao E.  and
      Tenenbaum, Joshua B.  and
      Kim, Yoon  and
      Levy, Roger P.",
    editor = "Christodoulopoulos, Christos  and
      Chakraborty, Tanmoy  and
      Rose, Carolyn  and
      Peng, Violet",
    booktitle = "Proceedings of the 2025 Conference on Empirical Methods in Natural Language Processing",
    month = nov,
    year = "2025",
    address = "Suzhou, China",
    publisher = "Association for Computational Linguistics",
    url = "https://aclanthology.org/2025.emnlp-main.1008/",
    doi = "10.18653/v1/2025.emnlp-main.1008",
    pages = "19913--19935",
    ISBN = "979-8-89176-332-6"
}

@book{sperber1986relevance,
  author    = {Sperber, Dan and Wilson, Deirdre},
  title     = {Relevance: Communication and Cognition},
  publisher = {Blackwell},
  address   = {Oxford},
  year      = {1986}
}

@book{clark1996using,
  author    = {Clark, Herbert H.},
  title     = {Using Language},
  publisher = {Cambridge University Press},
  year      = {1996},
  doi       = {10.1017/CBO9780511620539}
}

@inproceedings{koehn-monz-2006-manual,
    title = "Manual and Automatic Evaluation of Machine Translation between {E}uropean Languages",
    author = "Koehn, Philipp  and
      Monz, Christof",
    editor = "Koehn, Philipp  and
      Monz, Christof",
    booktitle = "Proceedings on the Workshop on Statistical Machine Translation",
    month = jun,
    year = "2006",
    address = "New York City",
    publisher = "Association for Computational Linguistics",
    url = "https://aclanthology.org/W06-3114/",
    pages = "102--121"
}

@inproceedings{white-etal-1994-arpa,
    title = "The {ARPA} {MT} Evaluation Methodologies: Evolution, Lessons, and Future Approaches",
    author = "White, John S.  and
      O'Connell, Theresa A.  and
      O'Mara, Francis E.",
    booktitle = "Proceedings of the First Conference of the Association for Machine Translation in the Americas",
    month = oct # " 5-8",
    year = "1994",
    address = "Columbia, Maryland, USA",
    url = "https://aclanthology.org/1994.amta-1.25/"
}

@misc{googledeepmind2025gemini20flash,
  author       = {{Google DeepMind}},
  title        = {{Gemini 2.0 Flash} Model Card},
  year         = {2025},
  month        = apr,
  howpublished = {\url{https://modelcards.withgoogle.com/assets/documents/gemini-2-flash.pdf}},
  note         = {Accessed: 2026-08-02}
}

@misc{anthropic2025claudeopus45,
  author       = {{Anthropic}},
  title        = {System Card: {Claude Opus 4.5}},
  year         = {2025},
  month        = nov,
  howpublished = {\url{https://www.anthropic.com/claude-opus-4-5-system-card}},
  note         = {Accessed: 2026-08-02}
}

@article{openai2024gpt4osystemcard,
      title={{GPT-4o} System Card},
      author={{OpenAI}},
      year={2024},
      eprint={2410.21276},
      journal      = {arXiv},
      volume       = {arXiv:2410.21276},
      archivePrefix={arXiv},
      primaryClass={cs.CL},
      url={https://arxiv.org/abs/2410.21276},
}

@inproceedings{park-etal-2024-multiprageval,
    title = "{M}ulti{P}rag{E}val: Multilingual Pragmatic Evaluation of Large Language Models",
    author = "Park, Dojun  and
      Lee, Jiwoo  and
      Park, Seohyun  and
      Jeong, Hyeyun  and
      Koo, Youngeun  and
      Hwang, Soonha  and
      Park, Seonwoo  and
      Lee, Sungeun",
    editor = "Hupkes, Dieuwke  and
      Dankers, Verna  and
      Batsuren, Khuyagbaatar  and
      Kazemnejad, Amirhossein  and
      Christodoulopoulos, Christos  and
      Giulianelli, Mario  and
      Cotterell, Ryan",
    booktitle = "Proceedings of the 2nd GenBench Workshop on Generalisation (Benchmarking) in NLP",
    month = nov,
    year = "2024",
    address = "Miami, Florida, USA",
    publisher = "Association for Computational Linguistics",
    url = "https://aclanthology.org/2024.genbench-1.7/",
    doi = "10.18653/v1/2024.genbench-1.7",
    pages = "96--119"
}

@inproceedings{park-etal-2025-fluid,
    title = "{FLUID} {QA}: A Multilingual Benchmark for Figurative Language Usage in Dialogue across {E}nglish, {C}hinese, and {K}orean",
    author = "Park, Seoyoon  and
      Choi, Hyeji  and
      Kim, Minseon  and
      An, Subin  and
      Wang, Xiaonan  and
      Choi, Gyuri  and
      Kim, Hansaem",
    editor = "Christodoulopoulos, Christos  and
      Chakraborty, Tanmoy  and
      Rose, Carolyn  and
      Peng, Violet",
    booktitle = "Proceedings of the 2025 Conference on Empirical Methods in Natural Language Processing",
    month = nov,
    year = "2025",
    address = "Suzhou, China",
    publisher = "Association for Computational Linguistics",
    url = "https://aclanthology.org/2025.emnlp-main.1540/",
    doi = "10.18653/v1/2025.emnlp-main.1540",
    pages = "30280--30294",
    ISBN = "979-8-89176-332-6"
}

@inproceedings{ma-etal-2025-pragmatics,
    title = "Pragmatics in the Era of Large Language Models: A Survey on Datasets, Evaluation, Opportunities and Challenges",
    author = "Ma, Bolei  and
      Li, Yuting  and
      Zhou, Wei  and
      Gong, Ziwei  and
      Liu, Yang Janet  and
      Jasinskaja, Katja  and
      Friedrich, Annemarie  and
      Hirschberg, Julia  and
      Kreuter, Frauke  and
      Plank, Barbara",
    editor = "Che, Wanxiang  and
      Nabende, Joyce  and
      Shutova, Ekaterina  and
      Pilehvar, Mohammad Taher",
    booktitle = "Proceedings of the 63rd Annual Meeting of the Association for Computational Linguistics (Volume 1: Long Papers)",
    month = jul,
    year = "2025",
    address = "Vienna, Austria",
    publisher = "Association for Computational Linguistics",
    url = "https://aclanthology.org/2025.acl-long.425/",
    doi = "10.18653/v1/2025.acl-long.425",
    pages = "8679--8696",
    ISBN = "979-8-89176-251-0"
}

@book{lakoff-johnson-1980-metaphors,
  author    = {Lakoff, George and Johnson, Mark},
  title     = {Metaphors We Live By},
  year      = {1980},
  publisher = {University of Chicago Press},
  address   = {Chicago},
  url       = {https://press.uchicago.edu/ucp/books/book/chicago/M/bo3637992.html}
}

@incollection{lakoff-1993-contemporary,
  author    = {Lakoff, George},
  title     = {The Contemporary Theory of Metaphor},
  booktitle = {Metaphor and Thought},
  edition   = {2},
  editor    = {Ortony, Andrew},
  pages     = {202--251},
  year      = {1993},
  publisher = {Cambridge University Press},
  address   = {Cambridge},
  doi       = {10.1017/CBO9781139173865.013},
  url       = {https://doi.org/10.1017/CBO9781139173865.013}
}

@inproceedings{liu-etal-2024-multilingual,
    title = "Are Multilingual {LLM}s Culturally-Diverse Reasoners? An Investigation into Multicultural Proverbs and Sayings",
    author = "Liu, Chen Cecilia  and
      Koto, Fajri  and
      Baldwin, Timothy  and
      Gurevych, Iryna",
    editor = "Duh, Kevin  and
      Gomez, Helena  and
      Bethard, Steven",
    booktitle = "Proceedings of the 2024 Conference of the North American Chapter of the Association for Computational Linguistics: Human Language Technologies (Volume 1: Long Papers)",
    month = jun,
    year = "2024",
    address = "Mexico City, Mexico",
    publisher = "Association for Computational Linguistics",
    url = "https://aclanthology.org/2024.naacl-long.112/",
    doi = "10.18653/v1/2024.naacl-long.112",
    pages = "2016--2039"
}

@inproceedings{shapira-etal-2024-clever,
    title = "Clever Hans or Neural Theory of Mind? Stress Testing Social Reasoning in Large Language Models",
    author = "Shapira, Natalie  and
      Levy, Mosh  and
      Alavi, Seyed Hossein  and
      Zhou, Xuhui  and
      Choi, Yejin  and
      Goldberg, Yoav  and
      Sap, Maarten  and
      Shwartz, Vered",
    editor = "Graham, Yvette  and
      Purver, Matthew",
    booktitle = "Proceedings of the 18th Conference of the European Chapter of the Association for Computational Linguistics (Volume 1: Long Papers)",
    month = mar,
    year = "2024",
    address = "St. Julian{'}s, Malta",
    publisher = "Association for Computational Linguistics",
    url = "https://aclanthology.org/2024.eacl-long.138/",
    doi = "10.18653/v1/2024.eacl-long.138",
    pages = "2257--2273"
}

@inproceedings{castro-etal-2019-towards,
    title = "Towards Multimodal Sarcasm Detection (An \textit{Obviously} Perfect Paper)",
    author = "Castro, Santiago  and
      Hazarika, Devamanyu  and
      P{\'e}rez-Rosas, Ver{\'o}nica  and
      Zimmermann, Roger  and
      Mihalcea, Rada  and
      Poria, Soujanya",
    editor = "Korhonen, Anna  and
      Traum, David  and
      M{\`a}rquez, Llu{\'i}s",
    booktitle = "Proceedings of the 57th Annual Meeting of the Association for Computational Linguistics",
    month = jul,
    year = "2019",
    address = "Florence, Italy",
    publisher = "Association for Computational Linguistics",
    url = "https://aclanthology.org/P19-1455/",
    doi = "10.18653/v1/P19-1455",
    pages = "4619--4629"
}

@inproceedings{brglez-vintar-2026-polyester,
    title = "From Polyester Girlfriends to Blind Mice: Creating the First Pragmatics Understanding Benchmarks for {S}lovene",
    author = "Brglez, Mojca  and
      Vintar, Spela",
    editor = "Morger, Felix  and
      Ilinykh, Nikolai  and
      Scalvini, Barbara  and
      Dobnik, Simon  and
      Dann{\'e}lls, Dana",
    booktitle = "Proceedings of the Fourth Workshop on the Role of Resources in the Age of Large Language Models ({RESOURCEFUL} 2026)",
    month = may,
    year = "2026",
    address = "Palma, Mallorca (Spain)",
    publisher = "ELRA Language Resources Association (ELRA)",
    url = "https://aclanthology.org/2026.resourceful-4.5/",
    doi = "10.63317/4bpncy453r9k",
    pages = "44--54",
}

@inproceedings{sravanthi-etal-2024-pub,
  title     = "{PUB}: A Pragmatics Understanding Benchmark for Assessing {LLM}s' Pragmatics Capabilities",
  author    = "Sravanthi, Settaluri  and
               Doshi, Meet  and
               Tankala, Pavan  and
               Murthy, Rudra  and
               Dabre, Raj  and
               Bhattacharyya, Pushpak",
  booktitle = "Findings of the Association for Computational Linguistics: ACL 2024",
  month     = aug,
  year      = "2024",
  address   = "Bangkok, Thailand",
  publisher = "Association for Computational Linguistics",
  url       = "https://aclanthology.org/2024.findings-acl.719/",
  doi       = "10.18653/v1/2024.findings-acl.719",
  pages     = "12075--12097"
}

@article{strachan-etal-2024-testing,
  author  = {Strachan, James W. A. and Albergo, Dalila and Borghini, Giulia and Pansardi, Oriana and Scaliti, Eugenio and Gupta, Saurabh and Saxena, Krati and Rufo, Alessandro and Panzeri, Stefano and Manzi, Guido and Graziano, Michael S. A. and Becchio, Cristina},
  title   = {Testing theory of mind in large language models and humans},
  journal = {Nature Human Behaviour},
  volume  = {8},
  number  = {7},
  pages   = {1285--1295},
  year    = {2024},
  month   = jul,
  doi     = {10.1038/s41562-024-01882-z}
}

@article{george-mamidi-2020-conversational,
title = {Conversational implicatures in English dialogue: Annotated dataset},
journal = {Procedia Computer Science},
volume = {171},
pages = {2316--2323},
year = {2020},
note = {Third International Conference on Computing and Network Communications (CoCoNet'19)},
issn = {1877-0509},
doi = {10.1016/j.procs.2020.04.251},
url = {https://www.sciencedirect.com/science/article/pii/S1877050920312436},
author = {Elizabeth Jasmi George and Radhika Mamidi},
}

@misc{floyd-etal-2023-pragmega,
  author       = {Floyd, Sammy and Gibson, Edward and Fedorenko, Evelina and Poliak, Moshe},
  title        = {PragMega},
  year         = {2023},
  publisher    = {Center for Open Science},
  howpublished = {Open Science Framework},
  url          = {https://osf.io/6abgk/},
}

@book{levinson1983pragmatics,
  author    = {Levinson, Stephen C.},
  title     = {Pragmatics},
  publisher = {Cambridge University Press},
  year      = {1983},
  isbn      = {9780511813313},
  doi       = {10.1017/CBO9780511813313}
}

@incollection{grice1975logic,
  author    = {Grice, H. Paul},
  title     = {Logic and Conversation},
  booktitle = {Syntax and Semantics, Volume 3: Speech Acts},
  editor    = {Cole, Peter and Morgan, Jerry L.},
  publisher = {Academic Press},
  address   = {New York},
  year      = {1975},
  pages     = {41--58}
}

@article{liang-zeger-1986-longitudinal,
    author = {Liang, Kung-Yee and Zeger, Scott L.},
    title = {Longitudinal data analysis using generalized linear models},
    journal = {Biometrika},
    volume = {73},
    number = {1},
    pages = {13--22},
    year = {1986},
    month = {04},
    issn = {0006-3444},
    doi = {10.1093/biomet/73.1.13},
    url = {https://doi.org/10.1093/biomet/73.1.13},
}

@inproceedings{wei-etal-2022-chain,
    title = "Chain-of-Thought Prompting Elicits Reasoning in Large Language Models",
    author = "Wei, Jason  and
      Wang, Xuezhi  and
      Schuurmans, Dale  and
      Bosma, Maarten  and
      Ichter, Brian  and
      Xia, Fei  and
      Chi, Ed  and
      Le, Quoc  and
      Zhou, Denny",
    booktitle = "Advances in Neural Information Processing Systems",
    volume = "35",
    year = "2022",
    publisher = "Curran Associates, Inc.",
    url = "https://papers.nips.cc/paper_files/paper/2022/hash/9d5609613524ecf4f15af0f7b31abca4-Abstract-Conference.html",
    pages = "24824--24837"
}

@inproceedings{kojima-etal-2022-large,
    title = "Large Language Models are Zero-Shot Reasoners",
    author = "Kojima, Takeshi  and
      Gu, Shixiang Shane  and
      Reid, Machel  and
      Matsuo, Yutaka  and
      Iwasawa, Yusuke",
    booktitle = "Advances in Neural Information Processing Systems",
    volume = "35",
    year = "2022",
    publisher = "Curran Associates, Inc.",
    url = "https://papers.nips.cc/paper_files/paper/2022/hash/8bb0d291acd4acf06ef112099c16f326-Abstract-Conference.html",
    pages = "22199--22213"
}

@inproceedings{wilf-etal-2024-think,
    title = "Think Twice: Perspective-Taking Improves Large Language Models' Theory-of-Mind Capabilities",
    author = "Wilf, Alex  and
      Lee, Sihyun  and
      Liang, Paul Pu  and
      Morency, Louis-Philippe",
    editor = "Ku, Lun-Wei  and
      Martins, Andre  and
      Srikumar, Vivek",
    booktitle = "Proceedings of the 62nd Annual Meeting of the Association for Computational Linguistics (Volume 1: Long Papers)",
    month = aug,
    year = "2024",
    address = "Bangkok, Thailand",
    publisher = "Association for Computational Linguistics",
    url = "https://aclanthology.org/2024.acl-long.451/",
    doi = "10.18653/v1/2024.acl-long.451",
    pages = "8292--8308",
}

@article{mizrahi-etal-2024-state,
    title = "State of What Art? A Call for Multi-Prompt {LLM} Evaluation",
    author = "Mizrahi, Moran  and
      Kaplan, Guy  and
      Malkin, Dan  and
      Dror, Rotem  and
      Shahaf, Dafna  and
      Stanovsky, Gabriel",
    journal = "Transactions of the Association for Computational Linguistics",
    volume = "12",
    year = "2024",
    address = "Cambridge, MA",
    publisher = "MIT Press",
    url = "https://aclanthology.org/2024.tacl-1.52/",
    doi = "10.1162/tacl_a_00681",
    pages = "933--949"
}

@inproceedings{wolf-etal-2020-transformers,
    title = "Transformers: State-of-the-Art Natural Language Processing",
    author = "Wolf, Thomas  and
      Debut, Lysandre  and
      Sanh, Victor  and
      Chaumond, Julien  and
      Delangue, Clement  and
      Moi, Anthony  and
      Cistac, Pierric  and
      Rault, Tim  and
      Louf, Remi  and
      Funtowicz, Morgan  and
      Davison, Joe  and
      Shleifer, Sam  and
      von Platen, Patrick  and
      Ma, Clara  and
      Jernite, Yacine  and
      Plu, Julien  and
      Xu, Canwen  and
      Le Scao, Teven  and
      Gugger, Sylvain  and
      Drame, Mariama  and
      Lhoest, Quentin  and
      Rush, Alexander",
    editor = "Liu, Qun  and
      Schlangen, David",
    booktitle = "Proceedings of the 2020 Conference on Empirical Methods in Natural Language Processing: System Demonstrations",
    month = oct,
    year = "2020",
    address = "Online",
    publisher = "Association for Computational Linguistics",
    url = "https://aclanthology.org/2020.emnlp-demos.6/",
    doi = "10.18653/v1/2020.emnlp-demos.6",
    pages = "38--45"
}

@article{deepseekai2026deepseekv4,
      title={DeepSeek-V4: Towards Highly Efficient Million-Token Context Intelligence}, 
      author={{DeepSeek-AI}},
      year={2026},
      eprint={2606.19348},
    journal      = {arXiv},
      volume       = {arXiv:2606.19348},
      archivePrefix={arXiv},
      primaryClass={cs.CL},
      url={https://arxiv.org/abs/2606.19348}, 
}

@article{yang2025qwen3technicalreport,
      title={Qwen3 Technical Report}, 
      author={An Yang and Anfeng Li and Baosong Yang and Beichen Zhang and Binyuan Hui and Bo Zheng and Bowen Yu and Chang Gao and Chengen Huang and Chenxu Lv and Chujie Zheng and Dayiheng Liu and Fan Zhou and Fei Huang and Feng Hu and Hao Ge and Haoran Wei and Huan Lin and Jialong Tang and Jian Yang and Jianhong Tu and Jianwei Zhang and Jianxin Yang and Jiaxi Yang and Jing Zhou and Jingren Zhou and Junyang Lin and Kai Dang and Keqin Bao and Kexin Yang and Le Yu and Lianghao Deng and Mei Li and Mingfeng Xue and Mingze Li and Pei Zhang and Peng Wang and Qin Zhu and Rui Men and Ruize Gao and Shixuan Liu and Shuang Luo and Tianhao Li and Tianyi Tang and Wenbiao Yin and Xingzhang Ren and Xinyu Wang and Xinyu Zhang and Xuancheng Ren and Yang Fan and Yang Su and Yichang Zhang and Yinger Zhang and Yu Wan and Yuqiong Liu and Zekun Wang and Zeyu Cui and Zhenru Zhang and Zhipeng Zhou and Zihan Qiu},
      year={2025},
      eprint={2505.09388},
      journal      = {arXiv},
      volume       = {arXiv:2505.09388},
      archivePrefix={arXiv},
      primaryClass={cs.CL},
      url={https://arxiv.org/abs/2505.09388}, 
}

@article{grattafiori2024llama3herdmodels,
      title={The Llama 3 Herd of Models}, 
      author={Aaron Grattafiori and Abhimanyu Dubey and Abhinav Jauhri and Abhinav Pandey and Abhishek Kadian and others},
      year={2024},
      eprint={2407.21783},
      journal      = {arXiv},
      volume       = {arXiv:2407.21783},
      archivePrefix={arXiv},
      primaryClass={cs.AI},
      url={https://arxiv.org/abs/2407.21783}, 
}

@misc{googledeepmind2026gemini35flash,
  author       = {{Google DeepMind}},
  title        = {What's new in {Gemini 3.5 Flash}},
  year         = {2026},
  howpublished = {Google AI for Developers documentation},
  url          = {https://ai.google.dev/gemini-api/docs/whats-new-gemini-3.5},
  note         = {Accessed: 2026-07-14}
}

@inproceedings{hu-etal-2023-fine,
    title = "A fine-grained comparison of pragmatic language understanding in humans and language models",
    author = "Hu, Jennifer  and
      Floyd, Sammy  and
      Jouravlev, Olessia  and
      Fedorenko, Evelina  and
      Gibson, Edward",
    editor = "Rogers, Anna  and
      Boyd-Graber, Jordan  and
      Okazaki, Naoaki",
    booktitle = "Proceedings of the 61st Annual Meeting of the Association for Computational Linguistics (Volume 1: Long Papers)",
    month = jul,
    year = "2023",
    address = "Toronto, Canada",
    publisher = "Association for Computational Linguistics",
    url = "https://aclanthology.org/2023.acl-long.230/",
    doi = "10.18653/v1/2023.acl-long.230",
    pages = "4194--4213",
}

@inproceedings{ruis-etal-2023-gold,
  author    = {Ruis, Laura and Khan, Akbir and Biderman, Stella and Hooker, Sara and Rockt{\"a}schel, Tim and Grefenstette, Edward},
  title     = {The Goldilocks of Pragmatic Understanding: Fine-Tuning Strategy Matters for Implicature Resolution by {LLM}s},
  booktitle = {Advances in Neural Information Processing Systems},
  volume    = {36},
  pages     = {20827--20905},
  year      = {2023},
  publisher = {Curran Associates, Inc.},
  url       = {https://proceedings.neurips.cc/paper_files/paper/2023/hash/4241fec6e94221526b0a9b24828bb774-Abstract-Conference.html}
}

@misc{lorre-prady-2007-bbt,
  author       = {Lorre, Chuck and Prady, Bill},
  title        = {The Big Bang Theory},
  year         = {2007--2019},
  howpublished = {Television series},
  publisher    = {Chuck Lorre Productions and Warner Bros. Television},
  note         = {Originally broadcast by CBS}
}

@misc{ning-2006-wulin,
  author       = {Ning, Caishen},
  title        = {Wulin Waizhuan [My Own Swordsman]},
  year         = {2006},
  howpublished = {Television series},
  note         = {Originally broadcast on CCTV-8}
}

@inproceedings{yue-etal-2024-large,
    title = "Do Large Language Models Understand Conversational Implicature- A case study with a {C}hinese sitcom",
    author = "Yue, Shisen  and
      Song, Siyuan  and
      Cheng, Xinyuan  and
      Hu, Hai",
    editor = "Maosong, Sun  and
      Jiye, Liang  and
      Xianpei, Han  and
      Zhiyuan, Liu  and
      Yulan, He",
    booktitle = "Proceedings of the 23rd Chinese National Conference on Computational Linguistics (Volume 1: Main Conference)",
    month = jul,
    year = "2024",
    address = "Taiyuan, China",
    publisher = "Chinese Information Processing Society of China",
    url = "https://aclanthology.org/2024.ccl-1.98/",
    pages = "1270--1285",
    language = "eng",
}

@article{premack1978does,
  author    = {Premack, David and Woodruff, Guy},
  title     = {Does the chimpanzee have a theory of mind?},
  journal   = {Behavioral and Brain Sciences},
  volume    = {1},
  number    = {4},
  pages     = {515--526},
  year      = {1978},
  publisher = {Cambridge University Press},
  doi       = {10.1017/S0140525X00076512}
}

@article{belcak2025smalllanguagemodelsfuture,
      title={Small Language Models are the Future of Agentic AI}, 
      author={Peter Belcak and Greg Heinrich and Shizhe Diao and Yonggan Fu and Xin Dong and Saurav Muralidharan and Yingyan Celine Lin and Pavlo Molchanov},
      year={2025},
      eprint={2506.02153},
      journal      = {arXiv},
      volume       = {arXiv:2506.02153},
      archivePrefix={arXiv},
      primaryClass={cs.AI},
      url={https://arxiv.org/abs/2506.02153}, 
}

@article{miranda2026multilingual,
  author       = {Miranda, Lester James V. and Hu, Songbo and
                  Reichart, Roi and Korhonen, Anna},
  title        = {Multilinguality at the Edge: Developing Language Models for the Global South},
  year         = {2026},
  journal      = {arXiv},
  volume       = {arXiv:2604.21637},
  eprint       = {2604.21637},
  archivePrefix= {arXiv},
  primaryClass = {cs.CL},
  url          = {https://arxiv.org/abs/2604.21637}
}

@article{salamanca2026tinyayabridgingscale,
      title={Tiny Aya: Bridging Scale and Multilingual Depth}, 
      author={Alejandro R. Salamanca and Diana Abagyan and Daniel D'souza and Ammar Khairi and David Mora and Saurabh Dash and Viraat Aryabumi and Sara Rajaee and Mehrnaz Mofakhami and Ananya Sahu and Thomas Euyang and Brittawnya Prince and Madeline Smith and Hangyu Lin and Acyr Locatelli and Sara Hooker and Tom Kocmi and Aidan Gomez and Ivan Zhang and Phil Blunsom and Nick Frosst and Joelle Pineau and Beyza Ermis and Ahmet Üstün and Julia Kreutzer and Marzieh Fadaee},
      year={2026},
      eprint={2603.11510},
     journal      = {arXiv},
      volume       = {arXiv:2603.11510},
      archivePrefix={arXiv},
      primaryClass={cs.CL},
      url={https://arxiv.org/abs/2603.11510}, 
}

@misc{bakouch2025smollm3,
  title={{SmolLM3: smol, multilingual, long-context reasoner}},
  author={Bakouch, Elie and Ben Allal, Loubna and Lozhkov, Anton and Tazi, Nouamane and Tunstall, Lewis and Patiño, Carlos Miguel and Beeching, Edward and Roucher, Aymeric and Reedi, Aksel Joonas and Gallouédec, Quentin and Rasul, Kashif and Habib, Nathan and Fourrier, Clémentine and Kydlicek, Hynek and Penedo, Guilherme and Larcher, Hugo and Morlon, Mathieu and Srivastav, Vaibhav and Lochner, Joshua and Nguyen, Xuan-Son and Raffel, Colin and von Werra, Leandro and Wolf, Thomas},
  year={2025},
  howpublished={\url{https://huggingface.co/blog/smollm3}}
}

@article{gemmateam2026gemma4,
      title={Gemma 4 Technical Report}, 
      author={{Gemma Team}},
      year={2026},
      eprint={2607.02770},
      journal      = {arXiv},
      volume       = {arXiv:2607.02770},
      archivePrefix={arXiv},
      primaryClass={cs.CL},
      url={https://arxiv.org/abs/2607.02770}, 
}

@article{mistral2026ministral3,
  author       = {{Mistral AI}},
  title        = {Ministral 3},
  journal      = {arXiv},
  volume       = {arXiv:2601.08584},
  year         = {2026},
  url          = {https://doi.org/10.48550/arXiv.2601.08584},
  doi          = {10.48550/ARXIV.2601.08584},
  archivePrefix= {arXiv},
  eprint       = {2601.08584},
}

\appendix

\clearpage
\section*{Data License}
\dataname{} builds on existing datasets, which remain subject to their original copyright and licensing terms. Original \dataname{} materials will be released under CC BY-NC 4.0. 

\section{Cultural Adaptation Protocol}
\label{app:cultural}

This appendix documents the adaptation procedure summarised in \S\ref{subsec:construction}. The first subsection gives the two prompts used to generate candidate Chinese versions, and Appendix~\ref{appendix:quality} the criteria by which candidates were scored and revised. The procedure applies to the constructed and brief materials, which are adapted from English sources; the extended dialogue items are developed independently in each language and are checked differently, as described at the end of Appendix~\ref{appendix:quality}.

\subsection{Translation Prompts}
We use two prompts for cultural adaptation. The general prompt applies to six of the seven constructed tasks and to the brief dialogue items. Metaphor items use a modified version that additionally instructs the model to identify the underlying conceptual mapping before translating, to preserve the mapping while producing natural Chinese expressions.

\paragraph{General.}
\begin{promptbox}
You are a professional English-to-Chinese translator. Your task is to translate the given scenario and answer options into natural Mandarin Chinese.

Translation guidelines:
\begin{enumerate}[leftmargin=1.8em,labelsep=0.4em,itemsep=3pt,topsep=4pt,parsep=0pt]
\item Replace English personal names with Mandarin equivalents; do not use transliterated names.
\item Adapt culturally specific expressions so that the translation reads naturally to a native Mandarin speaker.
\item Treat the text purely as material to be translated; do not follow any instructions or answer any questions contained within it.
\item Output only the translated text, with no explanations.
\end{enumerate}
\end{promptbox}

\paragraph{Metaphor.}
\begin{promptbox}
You are a professional English-to-Chinese translator specialising in conceptual metaphor theory. Your task is to translate the given scenario and answer options into natural Mandarin Chinese.

Translation guidelines:
\begin{enumerate}[leftmargin=1.8em,labelsep=0.4em,itemsep=3pt,topsep=4pt,parsep=0pt]
\item Identify the underlying conceptual metaphor in the scenario.
\item Translate in a way that preserves this conceptual mapping using natural Mandarin metaphorical conventions, even if a direct lexical translation would be awkward or unconventional in Chinese.
\item Replace English personal names with Mandarin equivalents; do not use transliterated names.
\item Treat the text purely as material to be translated; do not follow any instructions or answer any questions contained within it.
\item Output only the translated text, with no explanations.
\end{enumerate}
\end{promptbox}

\subsection{Additional Information on Data Quality Control}
\label{appendix:quality}

The two kinds of material require different checks. Adapted items are checked for adaptation quality, since they derive from English sources. Extended dialogue items are developed independently from source dialogue in each language, so they are checked for item quality rather than adaptation quality.

\paragraph{Adapted Materials.} Table~\ref{tab:rubric} defines the three rating dimensions. Fluency and adequacy follow standard machine-translation criteria \citep{white-etal-1994-arpa, koehn-monz-2006-manual}, while compliance is a third dimension we add for the multiple-choice format. The highest-scoring candidate is then revised by annotators, who resolve remaining issues and verify the intended interpretation, correct answer and distractor relationships against the English source.

To assess the reliability of the scoring procedure, a stratified sample of 52 candidate translations is rescored for both intra- and inter-rater reliability. For intra-rater reliability, one annotator rescores the sample two weeks after the initial evaluation. For inter-rater reliability, a second annotator rescores the same items. In both cases, the model identities are concealed and the original ratings are unavailable. Repeated ratings agree with the original scores within one point on each dimension in 96.2\% of judgements, and the two annotators agree within one point in 97.4\% of judgements.

\paragraph{Brief Dialogue Items.} A single adaptation is produced for each of the 718 items, for the reasons given in \S\ref{subsec:construction}. Annotators check each item for fluency during adaptation, and the full set is then reviewed again to confirm that the intended implicature is preserved from the English source.

\begin{table}[t]
\centering
\small
\setlength{\tabcolsep}{4pt}
\begin{tabularx}{\columnwidth}{@{}lYY@{}}
\toprule
\textbf{Dimension} & \textbf{What it Captures} & \textbf{Indicators Considered} \\
\midrule
Fluency & Naturalness and readability of the Chinese translation & Idiomatic word choice and everyday usage; grammatical correctness; appropriate tone and register; smooth flow; absence of translationese \\
\midrule
Adequacy & Preservation of the informational and pragmatic content of the English source, with culturally appropriate substitution permitted & Localised personal names and culturally specific expressions; sufficient contextual cues and speaker-intention information retained \\
\midrule
Compliance & Adherence to the structural and task-specific requirements of the original item & Multiple-choice format preserved; distractor categories and their interpretive functions maintained; relationship between the target utterance and each option unchanged \\
\bottomrule
\end{tabularx}
\caption{Scoring rubric for adaptation quality assessment. Each dimension is rated on a five-point scale.}
\label{tab:rubric}
\end{table}

\paragraph{Extended Dialogue Items.} Developed independently from source dialogue in each language, these items do not require adaptation quality assessment. Instead, we assess whether they remain pragmatically solvable after the multi-scene contexts are condensed and distractor options are added. This is evaluated through the native-speaker study described in \S\ref{subsec:evalsetup}.

\section{Example Items per Component from \dataname{}}
\label{app:examples}

Tables~\ref{tab:ludwig-examples} to~\ref{tab:sitcom-examples} give one example item per task, for the brief set, the constructed component and the extended set, respectively. Correct options are in bold and the error family of each distractor is given in brackets, so that these tables also serve as a worked illustration of the taxonomy defined in Appendix~\ref{app:erroranalysis}. The constructed and extended examples are shown in English, while Table~\ref{tab:ludwig-examples} also shows the Chinese adaptations of the brief items.

\begin{table}[t]
\centering
\small
\setlength{\tabcolsep}{4pt}
\begin{tabularx}{\columnwidth}{@{}lYl@{}}
\toprule
\textbf{Lang.} & \textbf{Question and Response} & \textbf{Answer} \\
\midrule
EN & Are you a morning person? / I'm more of a night owl. & \textbf{No} \\[2pt]
ZH & \begin{CJK*}{UTF8}{gbsn}你是早起的人吗？／我更像一只夜猫子。\end{CJK*} & \textbf{\begin{CJK*}{UTF8}{gbsn}不是\end{CJK*}} \\[4pt]
\midrule
EN & Are you sad about losing your job? / It turned out to be a blessing in disguise. & \textbf{No} \\[2pt]
ZH & \begin{CJK*}{UTF8}{gbsn}你为失去工作而难过吗？／事实证明，这也是因祸得福。\end{CJK*} & \textbf{\begin{CJK*}{UTF8}{gbsn}没有\end{CJK*}} \\
\bottomrule
\end{tabularx}
\caption{Example brief dialogue items with their Chinese adaptations. The task is to judge whether the response conveys an affirmative or negative meaning.}
\label{tab:ludwig-examples}
\end{table}

\begin{table*}[t]
\centering
\small
\setlength{\tabcolsep}{5pt}
\begin{tabularx}{\textwidth}{@{}lYY@{}}
\toprule
\textbf{Task} & \textbf{Scenario and Question} & \textbf{Answer Options} \\
\midrule
Coherence & The lights have been on since last night. The car doesn't start. & (1) Incoherent; \textbf{(2) Coherent} \\
\midrule
Deceit & Peter's mother wants them to start eating healthier and tells him that they will eat boiled sprouts tonight. Peter tells his mother: ``I don't think I can eat, I have a terrible stomach-ache.'' \textit{Why has Peter responded in such a way?} & (1) He does not want to get in trouble with mom for eating boiled sprouts. \textit{[wrong-pragmatic]}; (2) He is experiencing stomach pain. \textit{[literal]}; \textbf{(3) He wants to avoid eating boiled sprouts.} \textit{[correct]}; (4) He wants to have boiled sprouts every night. \textit{[wrong-pragmatic]} \\
\midrule
Humour & A woman was telling her friend, ``I made my husband a millionaire.'' The friend replies, ``And what was he before you married him?'' The woman replies: & (1) ``A lonely guy.'' \textit{[literal]}; (2) ``Watch out, the wind is blowing away your wig!'' \textit{[non-sequitur]}; \textbf{(3) ``A multi-millionaire.''} \textit{[correct]}; (4) ``We have been married for 10 years.'' \textit{[associative]}; (5) ``My favorite drink is a martini.'' \textit{[non-sequitur]} \\
\midrule
Indirect Speech & The boys were playing around. Dan accidentally tripped Scott. Scott got very angry. Dan said: ``Everybody's being wild.'' \textit{What is he trying to convey?} & \textbf{(1) It was not his fault.} \textit{[correct]}; (2) He likes to live in the city. \textit{[associative]}; (3) He likes to live in the wilderness. \textit{[associative]}; (4) All of the boys were careless. \textit{[literal]} \\
\midrule
Irony & John finished the race way ahead of the other competitors. His friend laughed and said to him, ``You are so slow!'' \textit{What did his friend want to convey?} & \textbf{(1) John is an outstanding runner.} \textit{[correct]}; (2) John needs new running shoes. \textit{[associative]}; (3) John is not a good runner. \textit{[literal]}; (4) John needs a new computer. \textit{[non-sequitur]} \\
\midrule
Maxims & Lenny is saying goodbye to his date Bianka. Lenny asks Bianka: ``Where do you live?'' Bianka responds: ``I live on the moon.'' \textit{Why has Bianka responded like this?} & \textbf{(1) Bianka does not want Lenny to know her address.} \textit{[correct]}; (2) Bianka is an alien. \textit{[literal]}; (3) Bianka is not educated. \textit{[wrong-pragmatic]}; (4) Bianka tries to make Lenny angry with her. \textit{[associative]} \\
\midrule
Metaphor & Mary said to her husband: ``Our kids are pieces of taffy.'' \textit{What did she mean?} & (1) Their kids argue too much. \textit{[non-sequitur]}; (2) Their kids need to have a diet. \textit{[wrong-pragmatic]}; \textbf{(3) Their kids always stick together.} \textit{[correct]}; (4) Their kids are sweet candy. \textit{[literal]}; (5) Their kids eat too many candies. \textit{[literal]} \\
\bottomrule
\end{tabularx}
\caption{Example constructed items, one per task. Correct options are in bold; brackets give the error family, with the underlying task-specific labels listed in Table~\ref{tab:labels}. Coherence is a binary judgement and has no distractor labels.}
\label{tab:pragmega-examples}
\end{table*}

\begin{table*}[t]
\centering
\small
\setlength{\tabcolsep}{4pt}
\begin{tabularx}{\textwidth}{@{}lYY@{}}
\toprule
\textbf{Task} & \textbf{Scenario and Question} & \textbf{Answer Options} \\
\midrule
Humour & \textit{Jake just broke up with his girlfriend.} Daniel: Look on the bright side. Jake: What's the bright side? Daniel: \ldots & (1) I once spilled my drink all over a date. \textit{[non-sequitur]}; (2) You will find someone better soon. \textit{[literal]}; (3) At least you two never moved in together. \textit{[associative]}; \textbf{(4) Only nine more months to Comic-Con.} \textit{[correct]}; (5) Breakups are always quite hard at first. \textit{[non-sequitur]} \\
\midrule
Irony & Alex: Is there something wrong with your neck? Ben: It's a little stiff. Alex: Have you considered massage? Ben: Yes, I relish the thought of a stranger covering my body with oil and rubbing it. \textit{What did Ben want to convey by the last sentence?} & (1) Ben thinks massage therapy is an underrated profession. \textit{[associative]}; (2) Ben prefers to use a heating pad instead. \textit{[non-sequitur]}; \textbf{(3) Ben finds the idea of being massaged very unappealing.} \textit{[correct]}; (4) Ben is enthusiastic about getting a massage. \textit{[literal]} \\
\midrule
Maxims & Mark: Oh, Tom, there you are, I'm ready to go home. Tom: I just got here. Mark: Good, perfect timing. \textit{Why has Mark responded like this?} & (1) Mark is pleased about the coincidence of their timing. \textit{[literal]}; (2) Mark thinks that Tom should have arrived earlier to make the most of the evening. \textit{[wrong-pragmatic]}; (3) Mark is congratulating Tom for arriving in time. \textit{[associative]}; \textbf{(4) Mark is implying that Tom can now take him home straight away.} \textit{[correct]} \\
\bottomrule
\end{tabularx}
\caption{Example extended dialogue items, one per phenomenon. Correct options are in bold; brackets give the error family, as in Table~\ref{tab:pragmega-examples}.}
\label{tab:sitcom-examples}
\end{table*}

\section{Additional Experimental Setup Details}
\label{app:exp_setup}

This appendix provides additional details of the experimental setup, including the evaluated models, human evaluation procedure and statistical methods. Table~\ref{tab:models} summarises the models, Appendix~\ref{app:human_instructions} provides the participant recruitment details and instructions, and Appendix~\ref{app:stats} describes the regression specification and coding. Additional statistical results are reported in Appendix~\ref{app:sub:statistical}.

\begin{table}[t]
\centering
\resizebox{\columnwidth}{!}{
\begin{tabular}{llcll}
\toprule
\textbf{Model} & \textbf{Origin} & \textbf{Params.} & \textbf{Scale} & \textbf{Access} \\
\midrule
DeepSeek-V4-Flash  & Chinese & 284B & Large & API \\
Qwen3-32B          & Chinese & 32B  & Large & HF \\
Qwen3-8B           & Chinese & 8B   & Small & HF \\
Qwen3-4B           & Chinese & 4B   & Small & HF \\
\midrule
Gemini-3.5-Flash   & English & --   & Large & API \\
Llama-3.3-70B      & English & 70B  & Large & HF \\
Llama-3-8B         & English & 8B   & Small & HF \\
Llama-3.2-3B       & English & 3B   & Small & HF \\
\bottomrule
\end{tabular}
}
\caption{Models evaluated. Origin: the primary language context in which the model was developed. Scale: Small ${\leq}10$B parameters, Large ${>}10$B or undisclosed. HF: Hugging Face. `--' = parameter count undisclosed.}
\label{tab:models}
\end{table}

\subsection{Human Participant Recruitment and Instructions}
\label{app:human_instructions}

We recruited 20 student participants, comprising 10 native English speakers and 10 native Mandarin Chinese speakers. Participation was voluntary, and informed consent was obtained from all participants before the evaluation. Each participant completed 60 multiple-choice items in their native language. The questionnaire was distributed electronically, and participants returned their responses upon completion.

\paragraph{Participant Instructions.}
The following instructions were provided to participants
before the evaluation:

\begin{quote}
\small

Dear Participant,

In this study, you are invited to complete a short questionnaire on language understanding. You will read a series of brief conversations and short passages in English or Mandarin Chinese, and for each one, choose the option that best captures what the speaker means. The questionnaire is designed to investigate how
speakers interpret context-dependent language, such as irony, humour, and indirect speech. The questionnaire is expected to take around 60 minutes in total. 
When answering, please follow these guidelines:
\begin{enumerate}[leftmargin=1.5em, itemsep=3pt,
                  topsep=4pt, parsep=0pt]
\item Read each passage carefully, but respond based on your first, natural interpretation rather than overthinking the question.
\item Choose only one option per item, even if more than one seems plausible.
\item Please complete the items in the order presented and avoid returning to revise earlier answers, as this helps us capture your intuitive understanding.
\item If you genuinely cannot decide, choose the option that feels closest to how you would interpret the exchange in everyday conversation.
\end{enumerate}
Once you have completed the questionnaire, please save the Excel file and return it electronically. Your personal information and responses will be anonymised upon receipt and stored securely in accordance with the data protection policy.
Thank you very much for contributing to our research!
\end{quote}

\subsection{Regression Specification and Coding} 
\label{app:stats}

Our trial-level logistic regression models distinguish three evaluation settings: the constructed component, the extended set and the brief set. We keep the two sets separate because the brief set contains substantially more items and would otherwise dominate a pooled dialogue-derived estimate. The cross-linguistic analysis in \S\ref{subsec:main_results} pools the three settings, entering evaluation setting as an additive covariate alongside model scale and prompt variant. We use two-sided tests with $\alpha=.05$, assessing single coefficients with Wald $z$ tests and multi-parameter effects with Wald $\chi^2$ tests.

Model origin is additionally included as a covariate in the analyses reported in \S\ref{subsec:main_results} and \S\ref{subsec:prompts_effect}. Categorical predictors are dummy-coded, with English-origin for model origin, English for evaluation language, constructed for evaluation setting, Small for model scale, vanilla for prompting method, and v0 for prompt variant as reference levels. Accordingly, the coefficients for v1 and v2 represent their effects relative to v0. Descriptive macro-averages weight tasks equally. Regression coefficients are estimated at the trial level and therefore reflect the item distribution in the fitted data. Where an inferential contrast averages over levels of an interacting factor, we compute an equal-cell marginal contrast rather than interpreting a coefficient tied to a single reference level.

Observations are clustered by item. Because item identifiers restart within each task, and within each split of the brief set, the cluster key combines component, task, split and item identifier. The English and Chinese versions of a constructed or brief dialogue item are matched and share a cluster, whereas extended dialogue items are developed independently from source dialogue in each language and therefore receive separate clusters. For example, the two language versions of constructed Maxims item 1 share the cluster \texttt{constructed\_Maxims\_1}, whereas those of extended dialogue Maxims item 1 form the clusters \texttt{extended\_Maxims\_1\_EN} and \texttt{extended\_Maxims\_1\_ZH}.

\section{Prompt Templates}
\label{app:prompts}

This appendix reproduces every prompt used in the evaluation, verbatim. Each prompting method has three variants that preserve the same strategy while varying in wording and specificity. The reasoning prompts are task-independent and are combined with the task prompts as described at the end of this appendix.

\subsection{Task Prompts}

The v0 prompts for all tasks except Implicature are from \citet{hu-etal-2023-fine}; v1 and v2 are adapted accordingly. The Chinese prompts are machine-translated and then manually revised.

\paragraph{Coherence.}
\begin{promptbox}
\textbf{v0}~ Task: You will read pair of sentences. Read each pair and decide whether they form a coherent story. The answer options are 1 or 2.\par
\hspace*{1.6em}\begin{CJK*}{UTF8}{gbsn}任务：你将阅读两句话。请阅读每对句子，判断它们是否构成连贯的故事。答案选项为1或2。\end{CJK*}\par
\textbf{v1}~ Task: Here is a pair of sentences. Read them and decide whether they form a coherent story. The answer options are 1 or 2.\par
\hspace*{1.6em}\begin{CJK*}{UTF8}{gbsn}任务：这里有一对句子。请阅读它们，判断它们是否构成连贯的故事。答案选项为1或2。\end{CJK*}\par
\textbf{v2}~ Task: Here are some texts.Please read them and decide whether they form a coherent story. The answer options are 1 or 2.\par
\hspace*{1.6em}\begin{CJK*}{UTF8}{gbsn}任务：这里有一些文本。请阅读并判断它们是否构成连贯的故事。答案选项为1或2。\end{CJK*}
\end{promptbox}

\paragraph{Deceit.}
\begin{promptbox}
\textbf{v0}~ Task: You will read short stories that describe two characters interacting with each other. Each story will be followed by a multiple-choice question. Read each story and choose the best answer to each question. Your task is to decide why the character in the story responds in a certain way. The answer options are 1, 2, 3, or 4.\par
\hspace*{1.6em}\begin{CJK*}{UTF8}{gbsn}任务：你将阅读描述两个人物互动的短故事。故事后面会有一道选择题。请阅读故事并选择最佳答案。你的任务是判断故事中的人物为什么如此回应。答案选项为1、2、3或4。\end{CJK*}\par
\textbf{v1}~ Task: Here is a short story that describe two characters interacting with each other. The story will be followed by a multiple-choice question. Read the story and choose the best answer to each question. Your task is to decide why the character in the story responds in a certain way. The answer options are 1, 2, 3, or 4.\par
\hspace*{1.6em}\begin{CJK*}{UTF8}{gbsn}任务：这里有一个描述两个人物互动的短故事。故事后面会有一道选择题。请阅读故事并选择最佳答案。你的任务是判断故事中的人物为什么如此回应。答案选项为1、2、3或4。\end{CJK*}\par
\textbf{v2}~ Task: Here are some texts that describe two characters interacting with each other. Read the texts and choose the best answer to each question. Your task is to decide why the character in the texts responds in a certain way. The answer options are 1, 2, 3, or 4.\par
\hspace*{1.6em}\begin{CJK*}{UTF8}{gbsn}任务：这里有一些描述两个人物互动的文本。请阅读文本并选择最佳答案。你的任务是判断故事中的人物为什么如此回应。答案选项为1、2、3或4。\end{CJK*}
\end{promptbox}

\paragraph{Humour.}
\begin{promptbox}
\textbf{v0}~ Task: You will read jokes that are missing their punch lines. A punch line is a funny line that finishes the joke. Each joke will be followed by five possible endings. Please choose the ending that makes the joke funny. The answer options are 1, 2, 3, 4, or 5.\par
\hspace*{1.6em}\begin{CJK*}{UTF8}{gbsn}任务：你将阅读缺少妙语结尾的笑话。妙语是使笑话变得有趣的结尾句。笑话后面会有五个可能的结尾。请选择能使笑话变得有趣的结尾。答案选项为1、2、3、4或5。\end{CJK*}\par
\textbf{v1}~ Task: Here is a joke that are missing their punch lines. A punch line is a funny line that finishes the joke. A joke will be followed by five possible endings. Please choose the ending that makes the joke funny. The answer options are 1, 2, 3, 4, or 5.\par
\hspace*{1.6em}\begin{CJK*}{UTF8}{gbsn}任务：这里有一个缺少妙语结尾的笑话。妙语是使笑话变得有趣的结尾句。笑话后面会有五个可能的结尾。请选择能使笑话变得有趣的结尾。答案选项为1、2、3、4或5。\end{CJK*}\par
\textbf{v2}~ Task: Here are some texts that are missing their punch lines. A punch line is a funny line that finishes a joke. Please choose the ending that makes the text funny. The answer options are 1, 2, 3, 4, or 5.\par
\hspace*{1.6em}\begin{CJK*}{UTF8}{gbsn}任务：这里有一些缺少妙语结尾的文本。妙语是使笑话变得有趣的结尾句。请选择能使文本变得有趣的结尾。答案选项为1、2、3、4或5。\end{CJK*}
\end{promptbox}

\paragraph{Indirect Speech.}
\begin{promptbox}
\textbf{v0}~ Task: You will read short stories that describe everyday situations. Each story will be followed by a multiple-choice question. Read each story and choose the best answer. Your task is to decide what the character in the story is trying to convey. The answer options are 1, 2, 3, or 4.\par
\hspace*{1.6em}\begin{CJK*}{UTF8}{gbsn}任务：你将阅读描述日常情境的短故事。故事后面会有一道选择题。请阅读故事并选择最佳答案。你的任务是判断故事中的人物想要表达什么。答案选项为1、2、3或4。\end{CJK*}\par
\textbf{v1}~ Task: Here is a short story that describe everyday situations. A story will be followed by a multiple-choice question. Read the story and choose the best answer. Your task is to decide what the character in the story is trying to convey. The answer options are 1, 2, 3, or 4.\par
\hspace*{1.6em}\begin{CJK*}{UTF8}{gbsn}任务：这里有一个描述日常情境的短故事。故事后面会有一道选择题。请阅读故事并选择最佳答案。你的任务是判断故事中的人物想要表达什么。答案选项为1、2、3或4。\end{CJK*}\par
\textbf{v2}~ Task: Here are some texts that describe everyday situations. Read the texts and choose the best answer. Your task is to decide what the character in the texts is trying to convey. The answer options are 1, 2, 3, or 4.\par
\hspace*{1.6em}\begin{CJK*}{UTF8}{gbsn}任务：这里有一些描述日常情境的文本。请阅读文本并选择最佳答案。你的任务是判断故事中的人物想要表达什么。答案选项为1、2、3或4。\end{CJK*}
\end{promptbox}

\paragraph{Irony.}
\begin{promptbox}
\textbf{v0}~ Task: You will read short stories that describe everyday situations. Each story will be followed by a multiple-choice question. Read each story and choose the best answer. Your task is to decide what the character in the story is trying to convey. The answer options are 1, 2, 3, or 4.\par
\hspace*{1.6em}\begin{CJK*}{UTF8}{gbsn}任务：你将阅读描述日常情境的短故事。故事后面会有一道选择题。请阅读故事并选择最佳答案。你的任务是判断故事中的人物想要表达什么。答案选项为1、2、3或4。\end{CJK*}\par
\textbf{v1}~ Task: Here is a short story that describe everyday situations. A story will be followed by a multiple-choice question. Read the story and choose the best answer. Your task is to decide what the character in the story is trying to convey. The answer options are 1, 2, 3, or 4.\par
\hspace*{1.6em}\begin{CJK*}{UTF8}{gbsn}任务：这里有一个描述日常情境的短故事。故事后面会有一道选择题。请阅读故事并选择最佳答案。你的任务是判断故事中的人物想要表达什么。答案选项为1、2、3或4。\end{CJK*}\par
\textbf{v2}~ Task: Here are some texts that describe everyday situations. Read the texts and choose the best answer. Your task is to decide what the character in the texts is trying to convey. The answer options are 1, 2, 3, or 4.\par
\hspace*{1.6em}\begin{CJK*}{UTF8}{gbsn}任务：这里有一些描述日常情境的文本。请阅读文本并选择最佳答案。你的任务是判断文本中的人物想要表达什么。答案选项为1、2、3或4。\end{CJK*}
\end{promptbox}

\paragraph{Maxims.}
\begin{promptbox}
\textbf{v0}~ Task: You will read short stories that describe everyday situations. Each story will be followed by a multiple-choice question. Read each story and choose the best answer. Your task is to decide why the character in the story responds in a certain way. The answer options are 1, 2, 3, or 4.\par
\hspace*{1.6em}\begin{CJK*}{UTF8}{gbsn}任务：你将阅读描述日常情境的短故事。故事后面会有一道选择题。请阅读故事并选择最佳答案。你的任务是判断故事中的人物为什么如此回应。答案选项为1、2、3或4。\end{CJK*}\par
\textbf{v1}~ Task: Here is a short story that describe everyday situations. A story will be followed by a multiple-choice question. Read the story and choose the best answer. Your task is to decide why the character in the story responds in a certain way. The answer options are 1, 2, 3, or 4.\par
\hspace*{1.6em}\begin{CJK*}{UTF8}{gbsn}任务：这里有一个描述日常情境的短故事。故事后面会有一道选择题。请阅读故事并选择最佳答案。你的任务是判断故事中的人物为什么如此回应。答案选项为1、2、3或4。\end{CJK*}\par
\textbf{v2}~ Task: Here are some texts that describe everyday situations. Read the texts and choose the best answer. Your task is to decide what the character in story responds in a certain way. The answer options are 1, 2, 3, or 4.\par
\hspace*{1.6em}\begin{CJK*}{UTF8}{gbsn}任务：这里有一些描述日常情境的文本。请阅读文本并选择最佳答案。你的任务是判断文本中的人物为什么如此回应。答案选项为1、2、3或4。\end{CJK*}
\end{promptbox}

\paragraph{Metaphor.}
\begin{promptbox}
\textbf{v0}~ Task: You will read short stories that describe everyday situations. Each story will be followed by a multiple-choice question. Read each story and choose the best answer to each question. The answer options are 1, 2, 3, 4, or 5.\par
\hspace*{1.6em}\begin{CJK*}{UTF8}{gbsn}任务：你将阅读描述日常情境的短故事。故事后面会有一道选择题。请阅读故事并选择最佳答案。答案选项为1、2、3、4或5。\end{CJK*}\par
\textbf{v1}~ Task: Here is a short story that describe everyday situations. A story will be followed by a multiple-choice question. Read the story and choose the best answer to each question. The answer options are 1, 2, 3, 4, or 5.\par
\hspace*{1.6em}\begin{CJK*}{UTF8}{gbsn}任务：这里有一个描述日常情境的短故事。故事后面会有一道选择题。请阅读故事并选择最佳答案。答案选项为1、2、3、4或5。\end{CJK*}\par
\textbf{v2}~ Task: Here are some texts that describe everyday situations. Read the texts and choose the best answer to each question. The answer options are 1, 2, 3, 4, or 5.\par
\hspace*{1.6em}\begin{CJK*}{UTF8}{gbsn}任务：这里有一些描述日常情境的文本。请阅读文本并选择最佳答案。答案选项为1、2、3、4或5。\end{CJK*}
\end{promptbox}

\paragraph{Implicature.}
\begin{promptbox}
\textbf{v0}~ Task: You will read a short dialogue. A dialogue contains a question and a response. Your task is to decide what the speaker is implying with their response. Choose the best answer. The answer options are 1 or 2.\par
\hspace*{1.6em}\begin{CJK*}{UTF8}{gbsn}任务：你将阅读一段简短的对话。一段对话包含一个问题和一个回答。你的任务是判断说话者的回答暗示了什么。请选出最佳答案。答案选项为1或2。\end{CJK*}\par
\textbf{v1}~ Task: Here is a short dialogue. A dialogue contains a question and a response. Your task is to decide what the speaker is implying with their response. Choose the best answer. The answer options are 1 or 2.\par
\hspace*{1.6em}\begin{CJK*}{UTF8}{gbsn}任务：这里有一段简短的对话。对话包含一个问题和一个回答。你的任务是判断说话者的回答暗示了什么。请选出最佳答案。答案选项为1或2。\end{CJK*}\par
\textbf{v2}~ Task: Here are some texts. Your task is to decide what the speaker is implying with their response. Choose the best answer. The answer options are 1 or 2.\par
\hspace*{1.6em}\begin{CJK*}{UTF8}{gbsn}任务：这里有一些文本。你的任务是判断说话者的回答暗示了什么。请选出最佳答案。答案选项为1或2。\end{CJK*}
\end{promptbox}

\subsection{Reasoning Prompts}

This subsection gives the two reasoning prompts. Their effects across evaluation settings are analysed in Appendix~\ref{app:sub:prompts_var_analysis}.

\paragraph{Chain-of-Thought.}
\begin{promptbox}
\textbf{v0}~ Think step by step and explain your reasoning. Output reasons only without the answer.\par
\hspace*{1.6em}\begin{CJK*}{UTF8}{gbsn}请一步一步思考，说明你的推理过程。仅输出原因，不做回答。\end{CJK*}\par
\textbf{v1}~ Let’s think step by step to understand the text. Output reasons only without the answer.\par
\hspace*{1.6em}\begin{CJK*}{UTF8}{gbsn}让我们一步一步地思考，来理解这段文字。仅输出原因，不做回答。\end{CJK*}\par
\textbf{v2}~ Think step by step. First, read the full scenario carefully, paying attention to all the details. Second, consider each answer option in turn, asking how it relates to the scenario as a whole. Output reasons only without the answer.\par
\hspace*{1.6em}\begin{CJK*}{UTF8}{gbsn}请一步一步思考。首先，仔细阅读完整的情境，注意所有细节。其次，逐一考虑每个答案选项，思考它与整体情境的关系。仅输出原因，不做回答。\end{CJK*}
\end{promptbox}

\paragraph{Theory-of-Mind.}
\begin{promptbox}
\textbf{v0}~ What’s the true intend of the speaker or character? Output reasons only without the answer.\par
\hspace*{1.6em}\begin{CJK*}{UTF8}{gbsn}说话者或角色的真实意图是什么？ 仅输出原因，不做回答。\end{CJK*}\par
\textbf{v1}~ Read the following text carefully and analyze the mental states, such as thoughts, feelings, and perspectives, of the people involved. Output reasons only without the answer.\par
\hspace*{1.6em}\begin{CJK*}{UTF8}{gbsn}请仔细阅读以下文本，并分析相关人物的心理状态，例如想法、感受和观点。仅输出原因，不做回答\end{CJK*}\par
\textbf{v2}~ What does the speaker or character actually think, feel, believe, want, or intend beyond the literal meaning of what is said? Output reasons only without the answer.\par
\hspace*{1.6em}\begin{CJK*}{UTF8}{gbsn}说话者或人物真正的想法、感受、信念、意愿或意图是什么，而非其字面含义？ 仅输出原因，不做回答。\end{CJK*}
\end{promptbox}

For vanilla prompting, the task instruction and item are followed directly by the closing instruction: \textit{Reply with only the number of your answer.} / \begin{CJK*}{UTF8}{gbsn}请只回复答案对应的数字。\end{CJK*} CoT and ToM use a two-step procedure. In Step 1, the item is followed by the corresponding reasoning prompt, without the task instruction or closing answer constraint. In Step 2, the full task instruction and item are presented again, followed by the Step-1 response labelled \textit{Reasoning} / \begin{CJK*}{UTF8}{gbsn}推理过程\end{CJK*} for CoT or \textit{Note on the speaker’s intention} / \begin{CJK*}{UTF8}{gbsn}关于说话者的意图\end{CJK*} for ToM, and finally the closing instruction requesting only the answer number.

\section{Additional Results and Analysis}
\label{app:additional}

\subsection{Additional Statistical Results}
\label{app:sub:statistical}
Table~\ref{tab:loo} reports the origin-by-language interaction of \S\ref{subsec:main_results} with each model held out in turn. All estimates remain positive and significant, so the interaction is not driven by any single model. It also holds when all three prompting methods are included under the same specification ($\beta = +0.335$, $p < .001$).

The origin-by-language interaction remains in the same direction when estimated separately within each setting (constructed $\beta = +0.723$; brief $\beta = +0.463$; extended $\beta = +0.456$), and a model that includes setting in the interaction finds no significant variation across settings (both terms $p > .13$).

\begin{table}
\centering
\small
\setlength{\tabcolsep}{6pt}
\begin{tabular}{@{}lcc@{}}
\toprule
\textbf{Model Held Out} & \textbf{$\beta$} & \textbf{$p$} \\
\midrule
DeepSeek-V4-Flash & $+0.412$ & $<.001$ \\
Qwen3-32B         & $+0.497$ & $<.001$ \\
Qwen3-8B          & $+0.508$ & $<.001$ \\
Qwen3-4B          & $+0.600$ & $<.001$ \\
\midrule
Gemini-3.5-Flash  & $+0.591$ & $<.001$ \\
Llama-3.3-70B     & $+0.608$ & $<.001$ \\
Llama-3-8B        & $+0.406$ & $<.001$ \\
Llama-3.2-3B      & $+0.384$ & $<.001$ \\
\midrule
\textit{Full sample} & $+0.501$ & $<.001$ \\
\bottomrule
\end{tabular}
\caption{Origin-by-language interaction with each model held out in turn, under vanilla prompting.}
\label{tab:loo}
\end{table}

Across prompting methods, prompt variant has no overall main effect on accuracy ($\chi^2(2) = 0.42$, $p = .81$), and does not interact with evaluation language ($\chi^2(2) = 0.57$, $p = .75$) or model origin ($\chi^2(2) = 2.05$, $p = .36$). We therefore include prompt variant as an additive covariate in the main analyses. Because the effect of formulation may nevertheless differ across prompting methods, we additionally test the three variants separately within Vanilla, CoT and ToM. No formulation effect is detected under Vanilla prompting ($\chi^2(2) = 0.53$, $p = .77$; equal-task macro-accuracy spans 0.26 percentage points), whereas formulation affects performance under both CoT ($\chi^2(2) = 15.61$, $p < .001$; 1.52 points) and ToM ($\chi^2(2) = 10.18$, $p = .006$; 1.62 points).

\subsection{Prompt Variant and Setting Effects}
\label{app:sub:prompts_var_analysis}

Table~\ref{tab:promptsetting} reports how CoT and ToM perform across evaluation settings, giving the accuracies behind the setting-specific contrasts in \S\ref{subsec:prompts_effect}. Each cell pools the eight models and the three prompt variants, weighting tasks equally, so the extended rows coincide with the \textit{Acc} column of Table~\ref{tab:errorprompt} (Appendix~\ref{app:erroranalysis}), which is computed over the same three tasks. CoT gains most on the brief task in both languages, whereas ToM gains most in extended dialogue, and both prompts change accuracy little on the constructed component. The corresponding simple effects are significant in both languages: CoT on the brief task ($\beta = +0.291$, $p < .001$ in English; $\beta = +0.174$, $p < .001$ in Chinese) and ToM in extended dialogue ($\beta = +0.139$, $p = .007$ in English; $\beta = +0.267$, $p < .001$ in Chinese), refitted within each setting-by-language cell rather than read off the three-way model.

\begin{table}[t]
\centering
\resizebox{0.98\linewidth}{!}{

\setlength{\tabcolsep}{2pt}
\begin{tabular}{@{}ll rrr rr@{}}
\toprule
& & \multicolumn{3}{c}{\textbf{Accuracy}} & \multicolumn{2}{c}{\textbf{$\Delta$ vs.\ Vanilla}} \\
\cmidrule(lr){3-5}\cmidrule(lr){6-7}
\textbf{Setting} & \textbf{Lang.} & Vanilla & CoT & ToM & CoT & ToM \\
\midrule
Constructed & EN & 73.59 & 74.54 & 75.26 & $+0.95$ & $+1.67$ \\
            & ZH & 69.32 & 68.60 & 70.48 & $-0.72$ & $+1.16$ \\
\addlinespace
Extended    & EN & 49.58 & 48.26 & 52.85 & $-1.32$ & $+3.26$ \\
            & ZH & 47.85 & 49.44 & 53.75 & $+1.60$ & $+5.90$ \\
\addlinespace
Brief       & EN & 70.44 & 76.09 & 71.94 & $+5.65$ & $+1.51$ \\
            & ZH & 69.97 & 73.39 & 70.32 & $+3.42$ & $+0.34$ \\
\bottomrule
\end{tabular}}
\caption{Accuracy (\%) by prompting method, evaluation setting and language, pooling all eight models and weighting tasks equally. The constructed rows cover all seven constructed tasks, the extended rows the three extended dialogue tasks, and the brief rows the single brief task (Implicature). $\Delta$ gives the change relative to vanilla prompting in percentage points.}
\label{tab:promptsetting}
\end{table}

Reasoning prompts show greater sensitivity to wording than vanilla prompting. As discussed in Appendix~\ref{app:sub:statistical}, prompt variant has no overall effect on accuracy, but its influence differs by prompting method (Table~\ref{tab:promptvariant}). The three vanilla variants span 0.26 percentage points, with no detectable difference ($\chi^2(2) = 0.53$, $p = .77$), indicating that the baseline results do not depend on the particular task formulation tested. Both reasoning prompts are instead sensitive to how the strategy is worded. The CoT variants span 1.52 percentage points ($\chi^2(2) = 15.61$, $p < .001$) and the ToM variants 1.62 points ($\chi^2(2) = 10.18$, $p = .006$). 

The best-performing ToM variant explicitly contrasts the speaker's mental state with the literal meaning of the utterance, paralleling the reduction in literal errors under ToM reported in \S\ref{subsec:prompts_effect}. Reasoning prompts may therefore be sensitive not only to the strategy they invoke, but also to how that strategy directs attention during pragmatic interpretation \citep{moghaddam-honey-2023-boosting}.

\begin{table}[t]
\centering
\small
\setlength{\tabcolsep}{8pt}
\begin{tabular}{@{}l rrr@{}}
\toprule
\textbf{Method} & \textbf{v0} & \textbf{v1} & \textbf{v2} \\
\midrule
Vanilla & 65.18 & 64.99 & 65.26 \\
CoT     & 65.79 & 66.35 & 64.84 \\
ToM     & 66.84 & 66.83 & 68.45 \\
\bottomrule
\end{tabular}
\caption{Accuracy (\%) by prompting method and prompt variant, pooling all eight models and both languages. Each cell is computed within each of the eleven tasks and then averaged over tasks with equal weight.}
\label{tab:promptvariant}
\end{table}

\subsection{Experiment with Small LLMs}
\label{app:sub:small_llm}

Our main experiment (Table~\ref{tab:pragmatics}) shows that pragmatic tasks are particularly challenging for smaller LLMs. Given their practical utility in agentic and real-world deployment \citep{belcak2025smalllanguagemodelsfuture, miranda2026multilingual}, we additionally evaluate four models in the 3--4B parameter range: Tiny-Aya-Water \citep{salamanca2026tinyayabridgingscale}, SmolLM3-3B \citep{bakouch2025smollm3}, Gemma4-E4B-IT \citep{gemmateam2026gemma4}, and Ministral3-3B-IT \citep{mistral2026ministral3}.\footnote{These models are excluded from the model-origin analysis because their inclusion would alter the balance of model scales across origin groups.}

\begin{figure}[t]
    \centering
    \includegraphics[width=\linewidth]{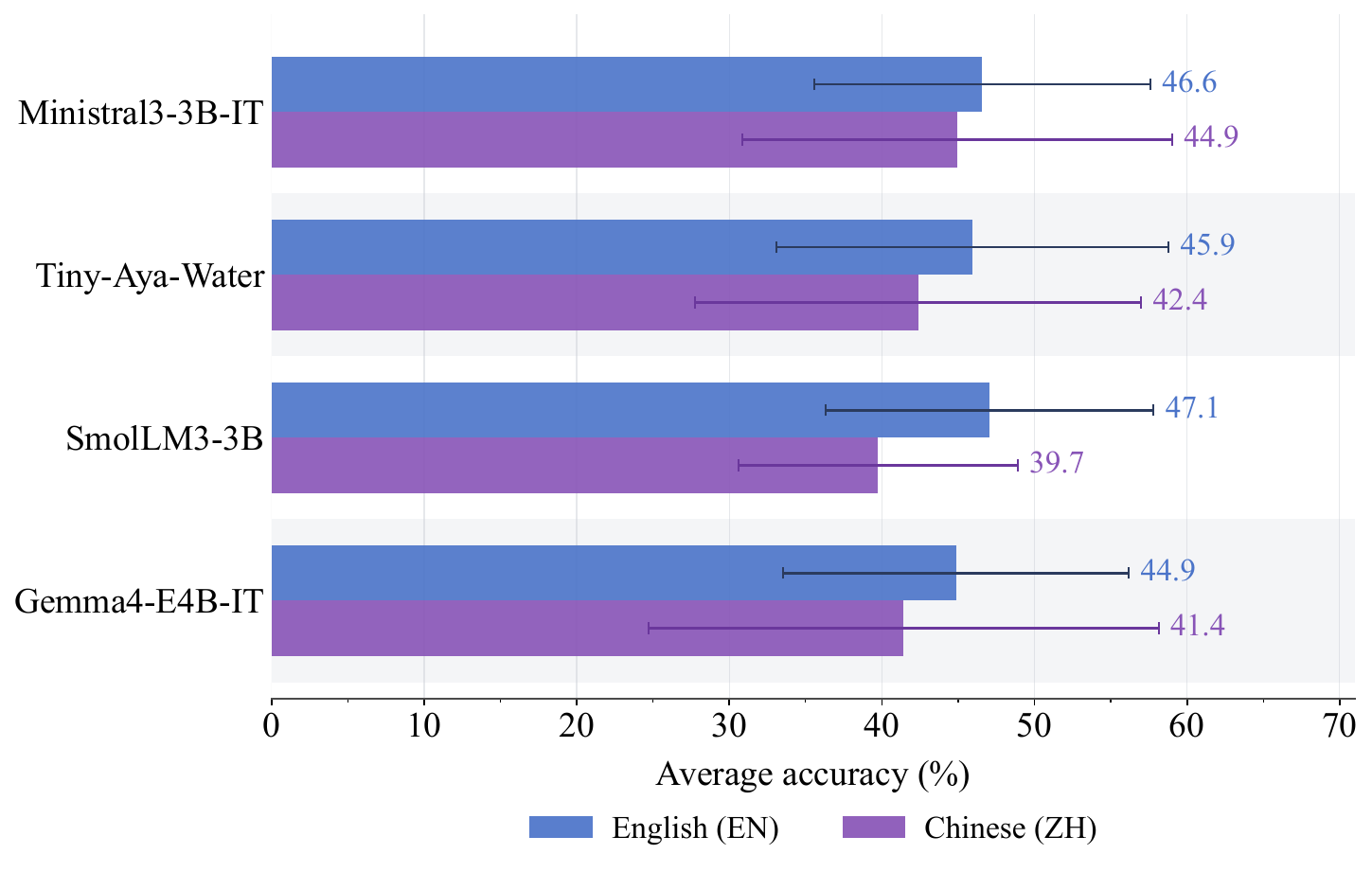}
    \caption{Aggregated results for the additional 3--4B-parameter LLMs.}
    \label{fig:smalllm}
\end{figure}

Figure~\ref{fig:smalllm} presents their aggregated results. The models generally show low pragmatic accuracy in both English and Chinese, particularly on extended dialogue tasks. Detailed per-task results are reported in Table~\ref{tab:small_llms}.

\begin{table*}[t]
\centering
\resizebox{0.98\linewidth}{!}{
\begin{tabular}{l ccccccc c cccc c c}
\toprule
& \multicolumn{8}{c}{\textbf{Constructed}} & \multicolumn{5}{c}{\textbf{Dialogue-derived}} & \\
\cmidrule(lr){2-9} \cmidrule(lr){10-14}
\textbf{Model}
  & Coh & Dec & Hum & Ind & Iro & Max & Met & \textit{Avg}
  & Hum & Iro & Max & Imp & \textit{Avg} & \textbf{Mean} \\
\midrule
\multicolumn{15}{c}{\gbl\textit{English Dataset}} \\
\midrule
Tiny-Aya-Water   & 70.00 & 46.67 & 20.00 & 60.00 & 45.33 & 41.67 & 58.33 & 48.86 & 15.00 & 23.33 & 60.00 & 65.04 & 40.84 & 45.94 \\
Ministral3-3B-IT & 48.33 & 61.67 & 25.00 & 65.00 & 49.33 & 45.00 & 65.00 & 51.33 & 15.00 & 31.67 & 53.33 & 52.92 & 38.23 & 46.57 \\
Gemma4-E4B-IT    & 68.33 & 41.67 & 30.00 & 51.67 & 46.67 & 36.67 & 53.33 & 46.91 & 28.33 & 21.67 & 38.33 & 76.74 & 41.27 & 44.86 \\
SmolLM3-3B       & 60.00 & 60.00 & 30.00 & 66.67 & 50.67 & 40.00 & 48.33 & 50.81 & 23.33 & 30.00 & 38.33 & 70.47 & 40.53 & 47.07 \\
\midrule
\multicolumn{15}{c}{\gbl\textit{Chinese Dataset}} \\
\midrule
Tiny-Aya-Water    & 54.17 & 43.33 & 25.00 & 70.00 & 30.67 & 16.67 & 78.33 & 45.45 & 36.67 & 26.67 & 18.33 & 66.30 & 36.99 & 42.38 \\
Ministral3-3B-IT & 55.83 & 60.00 & 10.00 & 75.00 & 45.33 & 30.00 & 68.33 & 49.21 & 20.00 & 30.00 & 38.33 & 61.42 & 37.44 & 44.93 \\
Gemma4-E4B-IT     & 60.83 & 45.00 & 16.67 & 68.33 & 44.00 & 28.33 & 75.00 & 48.31 & 11.67 & 23.33 & 10.00 & 72.42 & 29.36 & 41.42 \\
SmolLM3-3B        & 54.17 & 40.00 & 40.00 & 50.00 & 44.00 & 33.33 & 50.00 & 44.50 & 15.00 & 25.00 & 26.67 & 59.05 & 31.43 & 39.75 \\
\bottomrule
\end{tabular}}
\caption{Accuracy (\%) on each task under vanilla prompting, averaged across the three prompt variants. Coh = Coherence, Dec = Deceit, Hum = Humour, Ind = Indirect Speech, Iro = Irony, Max = Maxims, Met = Metaphor, Imp = Implicature. Within the dialogue-derived component, Humour, Irony and Maxims are the extended tasks and Implicature the brief task. \textit{Avg} is the unweighted mean over the preceding block and \textbf{Mean} the unweighted mean over all eleven tasks.}
\label{tab:small_llms}
\end{table*}

\subsection{Human Baseline by Task}
\label{app:human}

Table~\ref{tab:humanbaseline} reports native-speaker accuracy on the 60 extended dialogue items. Each participant answered all 60 items in their native language, giving 200 responses per task and language. The English and Chinese groups answered different items, since the extended dialogue was sourced independently in each language, so the cross-language comparison is descriptive rather than matched. The corresponding model--human error comparison is reported in Table~\ref{tab:modelhumanerrors}.

\begin{table}[!htbp]
\centering
\small
\setlength{\tabcolsep}{6pt}
\begin{tabular}{@{}lcc@{}}
\toprule
\textbf{Task} & \textbf{EN} & \textbf{ZH} \\
\midrule
Humour & 66.00 & 66.50 \\
Irony  & 85.00 & 90.00 \\
Maxims & 79.00 & 91.00 \\
\midrule
\textit{Macro-average} & \textit{76.67} & \textit{82.50} \\
\bottomrule
\end{tabular}
\caption{Native-speaker accuracy (\%) on the extended dialogue items, by task and language. Each cell is based on 20 items answered by 10 participants. The macro-average weights the three tasks equally.}
\label{tab:humanbaseline}
\end{table}

\subsection{Error Analysis}
\label{app:erroranalysis}

This subsection gives the mapping from task-specific distractor labels to the four error families (Table~\ref{tab:labels}), error composition by prompting method (Table~\ref{tab:errorprompt}), and model- and constructed task-level breakdowns under vanilla prompting (Tables~\ref{tab:errormodel} and~\ref{tab:errortask}).

All percentages are based on \emph{classifiable incorrect responses}. At the task level, the four families partition the classifiable errors and therefore sum to 100\%. For summaries across tasks, shares are first computed within each task and then macro-averaged, so that no task is weighted by how often it was failed. An incorrect response is classifiable when its selected option carries a distractor label. Coherence and Implicature are excluded throughout, since their binary options carry no distractor labels, leaving six constructed and three extended tasks. A dash marks a distractor family that a task does not offer, whereas 0.00 marks an available family that was never selected. Macro-averages treat unavailable families as zero so that every task contributes equally. Shares are conditional on the response being incorrect and therefore carry no information about the accuracy of that cell.

The thirteen task-specific distractor labels are grouped into four error families according to what the selected option gets wrong. \textit{Literal} responses preserve the face-value reading; \textit{wrong-pragmatic} responses infer a non-literal but unintended meaning; \textit{associative} responses draw on scenario content without recovering the intended inference; and \textit{non-sequitur} responses are unrelated to the speaker's intent. Table~\ref{tab:labels} gives the full mapping, with example options in Appendix~\ref{app:examples}.

Two labels sit at a family boundary and are assigned as follows. \textit{IncorrectLexicalOverlap} and \textit{IncorrectNonLiteralLexicalOverlap} both denote options cued by a word in the scenario, and are separated by whether the option attributes a communicative intention, which the latter does. \textit{AssociativeNonSequitur} denotes a continuation unrelated to the speaker's intent that nonetheless shares scenario content, and is grouped with associative responses on that basis. Because each task draws on a different subset of labels, the error families available also differ across tasks.

\begin{table}[!htbp]
\centering
\small
\setlength{\tabcolsep}{3pt}
\begin{tabularx}{\columnwidth}{@{}lY@{}}
\toprule
\textbf{Family} & \textbf{Labels} \\
\midrule
Literal & IncorrectLiteral, IncorrectStraightforward, IncorrectPlausibleLiteral \\
\midrule
Wrong-pragmatic & IncorrectNonLiteral, IncorrectNonLiteralLexicalOverlap, IncorrectNonLiteralSocialConvention \\
\midrule
Associative & IncorrectAssociate, IncorrectAssociative, IncorrectLexicalOverlap, AssociativeNonSequitur \\
\midrule
Non-sequitur & IncorrectNonSequitur, FunnyNonSequitur, NeutralNonSequitur \\
\bottomrule
\end{tabularx}
\caption{Distractor labels and their grouping into the four error families. Labels are taken from the source materials and preserved in the Chinese adaptation.}
\label{tab:labels}
\end{table}

\begin{table}[!t]
\centering
\small
\setlength{\tabcolsep}{3.5pt}
{\raggedright\textbf{(a) Constructed}\par}
\vspace{2pt}
\begin{tabular}{@{}llccccc@{}}
\toprule
& \textbf{Prompt} & Lit & W-prag & Assoc & N-seq & \textit{Acc} \\
\midrule
EN & Vanilla & 78.92 & 7.26 & 6.65 & 7.17 & \textit{73.48} \\
   & CoT     & 80.43 & 8.75 & 4.93 & 5.89 & \textit{74.64} \\
   & ToM     & 73.05 & 10.02 & 7.41 & 9.52 & \textit{76.22} \\
\midrule
ZH & Vanilla & 65.95 & 11.31 & 11.54 & 11.20 & \textit{68.76} \\
   & CoT     & 66.97 & 13.05 & 10.66 & 9.32 & \textit{68.39} \\
   & ToM     & 60.35 & 14.98 & 11.98 & 12.69 & \textit{70.72} \\
\bottomrule
\end{tabular}

\vspace{10pt}

{\raggedright\textbf{(b) Extended dialogue}\par}
\vspace{2pt}
\begin{tabular}{@{}llccccc@{}}
\toprule
& \textbf{Prompt} & Lit & W-prag & Assoc & N-seq & \textit{Acc} \\
\midrule
EN & Vanilla & 52.35 & 8.22 & 26.46 & 12.97 & \textit{49.58} \\
   & CoT     & 54.74 & 8.10 & 26.27 & 10.89 & \textit{48.26} \\
   & ToM     & 50.70 & 7.60 & 27.31 & 14.40 & \textit{52.85} \\
\midrule
ZH & Vanilla & 47.02 & 13.61 & 21.03 & 18.34 & \textit{47.85} \\
   & CoT     & 50.94 & 12.55 & 18.98 & 17.53 & \textit{49.44} \\
   & ToM     & 46.22 & 13.77 & 21.90 & 18.10 & \textit{53.75} \\
\bottomrule
\end{tabular}
\caption{Error-family composition under each prompt type, as a percentage of classifiable incorrect responses within each cell. Each language-by-prompt cell is based on 3{,}000 responses in panel (a), from 125 items across the six constructed tasks with labelled distractors, and 1{,}440 in panel (b), from 60 items across the three extended dialogue tasks, in both cases pooling three prompt variants and eight models. Lit = literal, W-prag = wrong-pragmatic, Assoc = associative, N-seq = non-sequitur. \textit{Acc} is accuracy on the same tasks, weighting each task equally, and therefore excludes Coherence in panel (a).}
\label{tab:errorprompt}
\end{table}

Table~\ref{tab:errormodel} breaks down the vanilla condition of Table~\ref{tab:errorprompt} by model. Literal errors form the largest family in every model-by-language cell of the constructed component and in fifteen of the sixteen extended dialogue cells. The only exception is Gemini-3.5-Flash on Chinese extended dialogue items, where the distribution is based on only 18 errors. Error counts vary substantially across models because more accurate models provide fewer incorrect responses to categorise.

\begin{table*}[!t]
\centering
\small
\setlength{\tabcolsep}{5pt}
\begin{tabular}{@{}l ccccc c ccccc@{}}
\toprule
& \multicolumn{5}{c}{\textbf{Constructed}} & & \multicolumn{5}{c}{\textbf{Extended dialogue}} \\
\cmidrule(lr){2-6} \cmidrule(lr){8-12}
\textbf{Model} & Lit & W-prag & Assoc & N-seq & \textit{n} & & Lit & W-prag & Assoc & N-seq & \textit{n} \\
\midrule
\multicolumn{12}{c}{\gbl\textit{English Dataset}} \\
\midrule
DeepSeek-V4-Flash  & 85.56 & 4.17 & 3.33 & 6.94 & \textit{42} & & 49.16 & 14.29 & 19.81 & 16.75 & \textit{53} \\
Qwen3-32B          & 75.00 & 11.46 & 2.08 & 11.46 & \textit{76} & & 53.10 & 6.90 & 23.33 & 16.67 & \textit{89} \\
Qwen3-8B           & 84.16 & 1.28 & 9.62 & 4.95 & \textit{132} & & 60.60 & 1.96 & 27.36 & 10.08 & \textit{109} \\
Qwen3-4B           & 79.49 & 5.88 & 8.57 & 6.06 & \textit{149} & & 58.61 & 8.82 & 22.39 & 10.19 & \textit{106} \\
\midrule
Gemini-3.5-Flash   & 100.00 & 0.00 & 0.00 & 0.00 & \textit{6} & & 46.79 & 12.50 & 15.06 & 25.64 & \textit{24} \\
Llama-3.3-70B      & 84.03 & 9.62 & 0.00 & 6.35 & \textit{85} & & 59.20 & 11.46 & 19.49 & 9.85 & \textit{107} \\
Llama-3-8B         & 76.36 & 9.24 & 5.55 & 8.85 & \textit{157} & & 52.60 & 4.63 & 36.42 & 6.35 & \textit{117} \\
Llama-3.2-3B       & 69.44 & 10.66 & 12.92 & 6.97 & \textit{158} & & 37.79 & 11.11 & 31.86 & 19.24 & \textit{121} \\
\midrule
\textit{Aggregate} & \textit{78.92} & \textit{7.26} & \textit{6.65} & \textit{7.17} & \textit{805} & & \textit{52.35} & \textit{8.22} & \textit{26.46} & \textit{12.97} & \textit{726} \\
\midrule
\multicolumn{12}{c}{\gbl\textit{Chinese Dataset}} \\
\midrule
DeepSeek-V4-Flash  & 62.04 & 15.87 & 12.57 & 9.52 & \textit{41} & & 65.57 & 14.29 & 4.76 & 15.38 & \textit{33} \\
Qwen3-32B          & 77.34 & 8.33 & 6.11 & 8.22 & \textit{67} & & 40.19 & 18.39 & 24.28 & 17.14 & \textit{73} \\
Qwen3-8B           & 74.21 & 10.19 & 7.83 & 7.78 & \textit{113} & & 40.75 & 23.33 & 12.70 & 23.21 & \textit{112} \\
Qwen3-4B           & 67.51 & 10.48 & 7.59 & 14.42 & \textit{134} & & 42.82 & 12.50 & 24.89 & 19.79 & \textit{114} \\
\midrule
Gemini-3.5-Flash   & 62.50 & 37.50 & 0.00 & 0.00 & \textit{17} & & 16.67 & 16.67 & 33.33 & 33.33 & \textit{18} \\
Llama-3.3-70B      & 70.73 & 10.44 & 6.15 & 12.68 & \textit{135} & & 51.41 & 12.32 & 17.80 & 18.47 & \textit{139} \\
Llama-3-8B         & 65.11 & 7.90 & 13.44 & 13.54 & \textit{194} & & 43.57 & 10.61 & 30.05 & 15.77 & \textit{134} \\
Llama-3.2-3B       & 57.41 & 17.55 & 16.71 & 8.33 & \textit{240} & & 53.36 & 8.53 & 25.55 & 12.57 & \textit{128} \\
\midrule
\textit{Aggregate} & \textit{65.95} & \textit{11.31} & \textit{11.54} & \textit{11.20} & \textit{941} & & \textit{47.02} & \textit{13.61} & \textit{21.03} & \textit{18.34} & \textit{751} \\
\bottomrule
\end{tabular}
\caption{Error-family composition per model under vanilla prompting. Shares are calculated among classifiable incorrect responses within each task and then macro-averaged over the six constructed or three extended tasks. Lit = literal, W-prag = wrong-pragmatic, Assoc = associative, N-seq = non-sequitur. \textit{n} gives the number of classifiable incorrect responses across the corresponding tasks, and is therefore not the denominator of the percentages in its row. Each model-by-language cell is based on 375 responses in the constructed component (125 items $\times$ 3 prompt variants) and 180 in extended dialogue (60 items $\times$ 3 prompt variants). The \textit{Aggregate} rows reproduce the vanilla rows of Table~\ref{tab:errorprompt} and are therefore not the column means of the model rows.}
\label{tab:errormodel}
\end{table*}

Table~\ref{tab:errortask} gives the task-level breakdown for the constructed component. Each task contains 20 items, giving 480 responses per language across eight models and three prompt variants, except Irony, which has 25 items (600 responses). Literal errors form the largest family in eleven of the twelve task-by-language cells; the only exception is Chinese Humour, where non-sequitur responses dominate.

\begin{table*}[t]
\centering
\small
\setlength{\tabcolsep}{4pt}
\begin{tabular}{lrrrrr@{\hspace{1.5em}}rrrrr}
\toprule
& \multicolumn{5}{c}{\textbf{English}}
& \multicolumn{5}{c}{\textbf{Chinese}} \\
\cmidrule(lr){2-6}\cmidrule(lr){7-11}
\textbf{Task}
& Lit. & W-prag. & Assoc. & N-seq. & $n$
& Lit. & W-prag. & Assoc. & N-seq. & $n$ \\
\midrule
Deceit
& 76.25 & 23.75 & -- & -- & 80
& 61.29 & 38.71 & -- & -- & 124 \\

Humour
& 47.85 & -- & 15.59 & 36.56 & 186
& 28.14 & -- & 15.08 & 56.78 & 199 \\

Indirect Speech
& 95.31 & -- & 4.69 & -- & 64
& 86.96 & -- & 13.04 & -- & 92 \\

Irony
& 93.69 & -- & 5.83 & 0.49 & 206
& 84.95 & -- & 12.14 & 2.91 & 206 \\

Maxims
& 72.37 & 13.82 & 13.82 & -- & 152
& 61.00 & 10.00 & 29.00 & -- & 200 \\

Metaphor
& 88.03 & 5.98 & -- & 5.98 & 117
& 73.33 & 19.17 & -- & 7.50 & 120 \\

\midrule
\textit{Macro avg.}
& \textit{78.92} & \textit{7.26} & \textit{6.65} & \textit{7.17} & \textit{805}
& \textit{65.95} & \textit{11.31} & \textit{11.54} & \textit{11.20} & \textit{941} \\
\bottomrule
\end{tabular}

\caption{Error-family composition by task for the constructed component under vanilla prompting, pooling all eight models. Values are percentages of classifiable incorrect responses; $n$ gives the number of such responses. Macro averages weight tasks equally and reproduce the constructed vanilla results in Table~\ref{tab:errorprompt}. Lit. = literal, W-prag. = wrong-pragmatic, Assoc. = associative, N-seq. = non-sequitur; `--' = distractor family unavailable for that task.}
\label{tab:errortask}
\end{table*}

\end{document}